\documentclass[letterpaper,twocolumn]{article}
\usepackage[margin=.8in]{geometry}
\usepackage[T1]{fontenc}
\usepackage{times}
\usepackage{natbib}
\usepackage{hyperref}
\usepackage{amsmath}
\usepackage{amssymb}
\usepackage{graphicx}
\usepackage{mleftright}
\usepackage{algorithm}
\usepackage[noend]{algpseudocode}

\DeclareMathOperator{\expect}{\mathbb{E}}
\usepackage{authblk}

\title{AlphaZeroES: Direct score maximization outperforms planning loss minimization}

\author[1]{Carlos Martin}
\author[1,2,3,4]{Tuomas Sandholm}
\affil[ ]{\{cgmartin, sandholm\}@cs.cmu.edu}
\affil[1]{Carnegie Mellon University}
\affil[2]{Strategy Robot, Inc.}
\affil[3]{Optimized Markets, Inc.}
\affil[4]{Strategic Machine, Inc.}
\date{}

\begin{document}
\maketitle
\begin{abstract}
Planning at execution time has been shown to dramatically improve performance for agents in both single-agent and multi-agent settings.
A well-known family of approaches to planning at execution time are AlphaZero and its variants, which use Monte Carlo Tree Search together with a neural network that guides the search by predicting state values and action probabilities.
AlphaZero trains these networks by minimizing a planning loss that makes the value prediction match the episode return, and the policy prediction at the root of the search tree match the output of the full tree expansion.
AlphaZero has been applied to both single-agent environments (such as Sokoban) and multi-agent environments (such as chess and Go) with great success.
In this paper, we explore an intriguing question:
In single-agent environments, can we outperform AlphaZero by directly maximizing the episode score instead of minimizing this planning loss, while leaving the MCTS algorithm and neural architecture unchanged?
To directly maximize the episode score, we use evolution strategies, a family of algorithms for zeroth-order blackbox optimization.
Our experiments indicate that, across multiple environments, directly maximizing the episode score outperforms minimizing the planning loss.
\end{abstract}

\section{Introduction}
\label{sec:introduction}

Lookahead search and reasoning is a central paradigm in artificial intelligence, and has a long history \citep{newell1965search,hart1968formal,nilsson1971problem,hart1972correction,lanctot2017unified,Brown18:Depth}.
In many domains, planning at execution time significantly improves performance.
In games like chess and Go, and single-agent domains like Sokoban, Pacman, and 2048, all state-of-the-art approaches use some form of planning by the agent.
Many planning approaches use \emph{Monte Carlo Tree Search (MCTS)}, which iteratively grows a search tree from the current state, and does so asymmetrically according to the information seen so far.
A prominent subfamily of approaches in this category are AlphaZero and its variants, which leverage function approximation via neural networks to learn good heuristic predictions of the values and action distributions at each state, which can be used to guide the tree search.
AlphaZero (and its variants) train this prediction function by minimizing a \emph{planning loss} consisting of a \emph{value loss} and a \emph{policy loss}.

In this paper, we set out to explore whether we can outperform AlphaZero and its variants in single-agent environments, where they are state of the art \citep{stochastic_muzero,gumbel_muzero}, by \emph{directly maximizing the episode score} instead, while leaving all other aspects of the agent, MCTS algorithm, and neural architecture unchanged.
Since MCTS is not differentiable, to maximize the episode score, we employ evolution strategies, a family of algorithms for zeroth-order blackbox optimization.

In \S\ref{sec:formulation}, we introduce our notation and present a detailed formulation of the problem.
In \S\ref{sec:related_research}, we describe prior related work and research.
In \S\ref{sec:method}, we present our method.
In \S\ref{sec:experiments}, we describe our experimental benchmarks and present our results.
In \S\ref{sec:conclusion}, we present our conclusion and suggest directions for future research.

\section{Problem formulation}
\label{sec:formulation}

In this section, we formulate the problem in detail and introduce some required notation.
If \(\mathcal{X}\) is a set, \(\triangle \mathcal{X}\) denotes the set of probability distributions on \(\mathcal{X}\).
An \emph{environment} is a tuple \((\mathcal{S}, \mathcal{A}, \rho, \delta)\) where \(\mathcal{S}\) is a set of states, \(\mathcal{A}\) is a set of actions, \(\rho : \triangle \mathcal{S}\) is an initial state distribution, and \(\delta : \mathcal{S} \times \mathcal{A} \to \mathbb{R} \times \mathbb{R} \times \mathcal{S}\) is a transition function.
A \emph{policy} is a function \(\mathcal{S} \to \triangle \mathcal{A}\) that maps a state to an action distribution.
Given an environment and policy, an \emph{episode} is a tuple \((s, a, r, \gamma)\) that is generated as follows.
First, an initial state \(s_0 \sim \rho\) is sampled.
Thereafter, on each timestep \(t \in \mathbb{N}\), an action \(a_t \sim \pi(s_t)\) is sampled, and a reward, discount factor, and new state \((r_t, \gamma_t, s_{t+1}) = \delta(s_t, a_t)\) are obtained.
The discount factor represents the probability of the episode ending at that timestep.
For a given episode, the \emph{return} at timestep \(t \in \mathbb{R}\) is defined recursively as \(R_t = r_t + \gamma_t R_{t+1}\).
The \emph{score} is the return at the initial timestep, \(R_0\).
An environment is said to \emph{terminate} at time \(t\) when \(\gamma_t = 0\).
The existence of such a \(t\) ensures that \(R_0\) is well-defined and finite.
Our goal is to find a policy \(\pi : \mathcal{S} \to \triangle \mathcal{A}\) that maximizes the expected score \(\expect R_0\).

\section{Related research}
\label{sec:related_research}

In this section, we describe related research.
Additional related research can be found in the appendix.

Monte Carlo methods are a wide class of computational algorithms that use repeated random sampling to estimate numerical quantities.
In the setting of planning, Monte-Carlo evaluation estimates the value of a position by averaging the return of several random rollouts.
\emph{Monte-Carlo Tree Search (MCTS)} \citep{mcts_original} combines Monte-Carlo evaluation with tree search.
Instead of backing-up the min-max value close to the root, and the average value at some depth, a more general backup operator is defined that progressively changes from averaging to min-max as the number of simulations grows.
This provides a fine-grained control of the tree growth and allows efficient selectivity.
A survey of recent modifications and applications of MCTS can be found in \citet{mcts_review}.

AlphaGo \citep{alphago} used a variant of MCTS to tackle the game of Go.
It used a neural network to evaluate board positions \emph{and} select moves.
These networks are trained using a combination of supervised learning from human expert games and reinforcement learning from self-play.
It was the first computer program to defeat a human professional player.
AlphaGo Zero \citep{alphago_zero} used reinforcement learning alone, \emph{without} any human data, guidance or domain knowledge beyond game rules.
AlphaZero \citep{alphazero} generalized AlphaGo Zero into a single algorithm that achieved superhuman performance in many challenging games, including chess and shogi.

MuZero~\citep{muzero} combined AlphaZero's tree-based search with a \emph{learned dynamics model}.
The latter allows it to plan in environments where the agent does \emph{not} have access to a simulator of the environment at execution time.
In the authors' words, ``All parameters of the model are trained jointly to accurately match the policy, value, and reward, for every hypothetical step \(k\), to corresponding target values observed after \(k\) actual time-steps have elapsed.''
Gumbel MuZero~\citep{gumbel_muzero} is a policy improvement algorithm based on sampling actions without replacement.
It replaces the more heuristic mechanisms by which AlphaZero selects and uses actions, both at root nodes and at non-root nodes.
It matches the state of the art on Go, chess, and Atari, and significantly improves prior performance when planning with few simulations.

\section{Proposed method}
\label{sec:method}

In this section, we present a detailed description of our proposed method, which we call AlphaZeroES.

\subsection{Planning algorithm}

MCTS grows a search tree asymmetrically, focusing on more promising subtrees.
Recent variants of MCTS, as described in Section~\ref{sec:related_research}, use value function approximation to guide the search.
For our experiments, we use the implementation of Gumbel MuZero \citep{gumbel_muzero}, which is the prior state of the art for this setting, found in the open-source Google JAX library mctx \citep{deepmind2020jax}.
MCTS iteratively constructs a search tree starting from some given state \(s_0\).
Each node in this tree contains a state, predicted value, predicted action probabilities, and, for each action, a visit count \(N\), action value \(Q\), reward, and discount factor.
Each iteration of the algorithm consists of three phases: \emph{selection}, \emph{expansion}, and \emph{backpropagation}.

During \emph{selection}, we start at the root and traverse the tree until a leaf edge is reached.
At internal nodes, we select actions according to the policy described in \citet{gumbel_muzero}.
When we reach a leaf edge \((s, a)\), we perform \emph{expansion} as follows.
We compute \((r, \gamma, s') = \delta(s, a)\), storing \(r\) and \(\gamma\) in the edge's parent node.
We then query the agent's \emph{prediction function} \((v, p) = f_\theta(s')\) to obtain the predicted value and action probabilities of \(s'\).
A new node is added to the tree containing this information, with action visit counts and action values initialized to zero.
Finally, we perform \emph{backpropagation} as follows.
The new node's value estimate is backpropagated up the tree to the root in the form of an \(n\)-step return.
Specifically, from \(t = T\) to \(0\), where \(T\) is the length of the trajectory, we compute an estimate of the cumulative discounted reward that bootstraps from the value estimate \(v_T\):
\(G_t = \sum_{t=0}^{T-1-t} \gamma^t r_{t+1+t} + \gamma^{T-t} v_T\).
For each such \(t\), we update the statistics for the edge \((s_t, a_t)\) as follows:
\(Q(s_t, a_t) \gets \frac{N(s_t, a_t) Q(s_t, a_t) + G_t}{N(s_t, a_t) + 1}\),
\(N(s_t, a_t) \gets N(s_t, a_t) + 1\).
The \emph{simulation budget} is the total number of iterations, which is the number of times the search tree is expanded, and therefore the size of the tree.

\subsection{Prediction function}

The prediction function of the agent takes an environment state as input and outputs a probability distribution over actions and value estimate.
Our experimental settings have states that are naturally modeled as \emph{sets} of objects (such as sets of cities, facilities, targets, boxes, \emph{etc.}), where each object can be described by a vector (\emph{e.g.}, the coordinates of a city and whether it has been visited or not).
Therefore, we seek a neural network architecture that can process a \emph{set} of vectors, rather than just a single vector.
Some early work on neural networks that can process sets was carried out by \citet{mcgregor2007neural} and \citet{mcgregor2008further}.

In our experiments, we use \emph{DeepSets \citep{zaheer2017deep}}, a neural network architecture that can process sets of inputs in a way that is equivariant or invariant (depending on the desired type of output) with respect to the inputs.
It is known to be a universal approximator for continuous set functions, provided that the model's latent space is sufficiently high-dimensional \citet{wagstaff2022universal}.
DeepSets may be viewed as the most efficient incarnation of the Janossy pooling paradigm \citep{murphy2018janossy}, and can be generalized by Transformers \citep{vaswani2017attention,kim2021transformers}.
A permutation-\emph{equi}variant layer of the DeepSets architecture has the form \citep[Supplement p. 19]{zaheer2017deep}
\(\mathbf{Y} = \sigma(\mathbf{X} \cdot \mathbf{A} + \mathbf{1} \otimes \mathbf{b} + \mathbf{1} \otimes ((\mathbf{1} \cdot \mathbf{X}) \cdot \mathbf{C}))\)
where
\(\mathbf{X} \in \mathbb{R}^{n \times d}\),
\(\mathbf{Y} \in \mathbb{R}^{n \times k}\),
\(\mathbf{A}, \mathbf{C} \in \mathbb{R}^{n \times k}\),
\(\mathbf{b} \in \mathbb{R}^k\),
and \(\mathbf{1}\) is the all-ones vector of appropriate dimensionality, and \(\sigma\) is a nonlinear activation function, such as ReLU.
A permutation-\emph{in}variant layer is simply a permutation-equivariant layer followed by global average pooling (yielding an output that is a vector rather than a matrix) followed by a nonlinearity.
In problems where the action space matches the set of inputs (such as cities in the TSP problem, or points in the vertex \(k\)-center and maximum diversity problems), the predicted action logits are read out via an affine dense layer following the permutation-equivariant layer, before global pooling.
In problems where the action space is a fixed set of actions (such as Sokoban and the navigation problems), the predicted action logits are read out via an affine dense layer following the permutation-invariant layer.
In both cases, the predicted value is read out via an affine dense layer from the output of the permutation-invariant layer.

\subsection{Training procedure}

AlphaZero minimizes a \emph{planning loss}, which is the sum of a \emph{value loss} and a \emph{policy loss}:
\(\mathcal{L} = \mathcal{L}_\text{value} + \mathcal{L}_\text{policy} = \sum_t (R_t - v_t)^2 + \textstyle\sum_t \operatorname{H}(w_t, p_t)\).
Here, \((v_t, p_t) = f_\theta(s_t)\) is the predicted state value and action probabilities for \(s_t\), respectively.
\((R_t - v_t)^2\) is the squared difference between \(v_t\) and the actual episode return \(R_t\).
\(\operatorname{H}(w_t, p_t)\) is the cross entropy between the action weights \(w_t\) returned by the MCTS algorithm for \(s_t\) and \(p_t\).
In contrast, in our approach, we fix \emph{the exact same architecture, hyperparameters, and MCTS algorithm}, but change the optimization objective.
Instead of minimizing the planning loss, we seek to \emph{maximize the episode score directly}.
One potential way we could seek to do this is by using policy gradient methods, which yield an estimator of the gradient of the expected return with respect to the agent's parameters.
There is a vast literature on policy gradient methods, which include REINFORCE \citep{reinforce} and actor-critic methods \citep{actor_critic_1,actor_critic_2}.

However, there is a problem.
Most of these methods assume that the policy is \emph{differentiable}---more precisely, that its output action distribution is differentiable with respect to the parameters of the policy.
However, our planning policy uses MCTS as a subroutine, and standard MCTS is not differentiable.
Because our policy contains a non-differentiable submodule, we need to find an alternative way to optimize the policy's parameters.
Furthermore, \citet{metz2021gradients} show that differentiation can be fail to be useful when trying to optimize certain functions---specifically, when working with an iterative differentiable system with chaotic dynamics.
Fortunately, we can turn to black-box (\emph{i.e.}, zeroth-order) optimization. Black-box optimization uses only function evaluations to optimize a black-box function with respect to a set of inputs.
In particular, it does not require gradients.
In our case, the black-box function maps our policy's parameters to a sampled episode score.

There is a class of black-box optimization algorithms called evolution strategies (ES)~\citep{Rechenberg_1973, Schwefel_1977, Rechenberg_1978} that maintain and evolve a population of parameter vectors. Natural evolution strategies (NES)~\citep{Wierstra_2014, Sun_2009} represent the population as a distribution over parameters and maximize its average objective value using the score function estimator.
For many parameter distributions, such as Gaussian smoothing, this is equivalent to evaluating the function at randomly-sampled points and estimating the gradient as a sum of estimates of directional derivatives along random directions \citep{Duchi_2015, Nesterov_2017, Shamir_2017, Berahas_2022}.

ES is a scalable alternative to standard reinforcement learning \citep{openai_es}, and also a viable method for learning non-differentiable parameters of large supervised models \citep{lenc2019non}.
We use OpenAI-ES~\citep{openai_es}, an NES algorithm that is based on the identity
\(\nabla_\mathbf{x} \expect_{\mathbf{z} \sim \mathcal{N}} f(\mathbf{x} + \sigma \mathbf{z}) = \tfrac{1}{\sigma} \expect_{\mathbf{z} \sim \mathcal{N}} f(\mathbf{x} + \sigma \mathbf{z}) \mathbf{z}\),
where \(\mathcal{N}\) is the standard multivariate normal distribution with the same dimension as \(\mathbf{x}\).

This algorithm works as follows.
Let \(\mathcal{I}\) be a set of indices.
For each \(i \in \mathcal{I}\) in parallel, sample \(\mathbf{z}_i \sim \mathcal{N}\) and compute \(\delta_i = f(\mathbf{x} + \sigma \mathbf{z}_i)\).
Finally, compute the pseudogradient \(\mathbf{g} = \frac{1}{\sigma |\mathcal{I}|} \sum_{i \in \mathcal{I}} \delta_i \mathbf{z}_i\).
To reduce variance, like \citet{openai_es}, we use antithetic sampling~\citep{Geweke_1988}, also called mirrored sampling~\citep{brockhoff2010mirrored}, which uses pairs of perturbations \(\pm \sigma \mathbf{z}_i\).
The resulting gradient is fed into an optimizer.
OpenAI-ES is massively parallelizable, since each \(\delta_i\) can be evaluated on a separate worker.
Furthermore, the entire optimization procedure can be performed with minimal communication bandwidth between workers.
All workers are initialized with the same random seed.
Worker \(i\) evaluates \(\delta_i\), sends it to the remaining workers, and receives the other workers' values (this is called an allgather operation in distributed computing).
Thus the workers compute the same \(\mathbf{g}\) and stay synchronized.
This process is described in Algorithm 2 of \citet{openai_es}.
The full training process is summarized in Algorithm 1.

\begin{algorithm}[t]
\caption{
Distributed training.
\\
The following algorithm runs on each worker.
All workers are initialized with the same random seed.
\\
The function \(f\) samples an episode given parameters for the agent, and outputs the episode score.
}
\label{alg:algorithm}
\begin{algorithmic}
\State \(\sigma \in \mathbb{R}\) is the perturbation scale
\State \(\mathbf{x} \in \mathbb{R}^d \gets \text{agent.initialize\_parameters}()\)
\State \(S \gets \text{optimizer.initialize\_state}(\mathbf{x})\)
\Loop
\State \(\mathbf{x} \gets \text{optimizer.get\_parameters}(S)\)
\State \(\mathcal{I} \gets\) set of available workers
\For{\(i \in \mathcal{I}\)}
\State \(\mathbf{z}_i \sim \mathcal{N}(\mathbf{0}_d, \mathbf{I}_d)\)
\EndFor
\State \(j \gets \) own worker rank
\State \(\delta_{j} \gets f(\mathbf{x} + \sigma \mathbf{z}_{j}) - f(\mathbf{x} - \sigma \mathbf{z}_{j})\)
\State send \(\delta_{j}\) to other workers
\State receive \(\{\delta_i\}_{i \in \mathcal{I} - \{j\}}\) from other workers
\State \(\mathbf{g} \gets \frac{1}{2 \sigma |\mathcal{I}|} \sum_{i \in \mathcal{I}} \delta_i \mathbf{z}_i\)
\State \(S \gets \text{optimizer.update\_state}(S, \mathbf{g})\)
\EndLoop
\end{algorithmic}
\end{algorithm}

\section{Experiments}
\label{sec:experiments}

In this section, we describe our experiments and present our results.
Unless stated otherwise, we use the following hyperparameters.
We use
10 trials per experiment,
training and evaluation batch sizes of 10,000,
1,000 batches per epoch,
4 hours of training time per trial,
the Adabelief \citep{zhuang2020adabelief} optimizer,
a perturbation scale of 0.1 for OpenAI-ES,
an MCTS simulation budget of 8 (Gumbel Muzero, the AlphaZero variant we use, can learn reliably with as few as 2 simulations \citep[p. 8]{gumbel_muzero}),
hidden layer sizes of 16 for the DeepSets network,
1 equivariant plus 1 invariant hidden layer for the DeepSets network,
and the ReLU activation function.

In our plots, we show the episode scores attained by AlphaZero (labeled \texttt{es=0} in the plot legend) vs. AlphaZeroES (labeled \texttt{es=1} in the plot legend).
To perform a fair comparison, since AlphaZero and AlphaZeroES optimize different objectives, we test both across a wide range of learning rates (labeled \texttt{lr} in the plot legend).
In addition, we show the value losses and policy losses over the course of training.
Though AlphaZeroES does not optimize these losses directly, we are interested in observing what happens to them as a side-effect of AlphaZeroES maximizing the episode score.
Solid lines show the mean across trials, and bands show the standard error of the mean.
We emphasize that it is not our goal to develop the best special-purpose solver for any one of these domains.
That is beyond the scope of this paper.
Rather, we are interested in a \emph{general}-purpose approach that can tackle \emph{all} of these domains, and learn good heuristics on its own by discovering generalizable patterns in data.
More narrowly, we wish to study how well an agent like AlphaZero can learn to leverage MCTS via direct score maximization rather than planning loss minimization.

\paragraph{Navigation problem}

In this environment, an agent navigates a gridworld to reach as many targets as possible within a given time limit.
At the beginning of each episode, targets are placed uniformly at random in a \(10 \times 10\) grid, as is the agent.
On each timestep, the agent can move up, down, left, or right by one tile.
The agent reaches a target when it moves into the same tile.
The agent receives a reward of \(+1\) when it reaches a target.
Thus the agent is incentivized to reach as many targets as possible within the time limit.
For our experiments, we use 20 targets and a time limit of 50 steps.
The prediction network observes a set of vectors, one for each target, where each vector contains the coordinates of the target, a boolean 0-1 flag indicating whether it has already been reached, and the number of episode timesteps remaining.

This environment has been used before as a benchmark by \citet[\S 4.2]{vpn}.
It resembles a traveling salesman-like problem in which several ``micro'' actions are required to perform the ``macro'' actions of moving from one city to another.
(Also, the agent can visit cities multiple times and does not need to return to its starting city.)
This models situations where several fine-grained actions are required to perform relevant tasks, such as moving a unit in a real-time strategy game a large distance across the map.

\begin{figure}
\centering
\includegraphics[width=.6\linewidth]{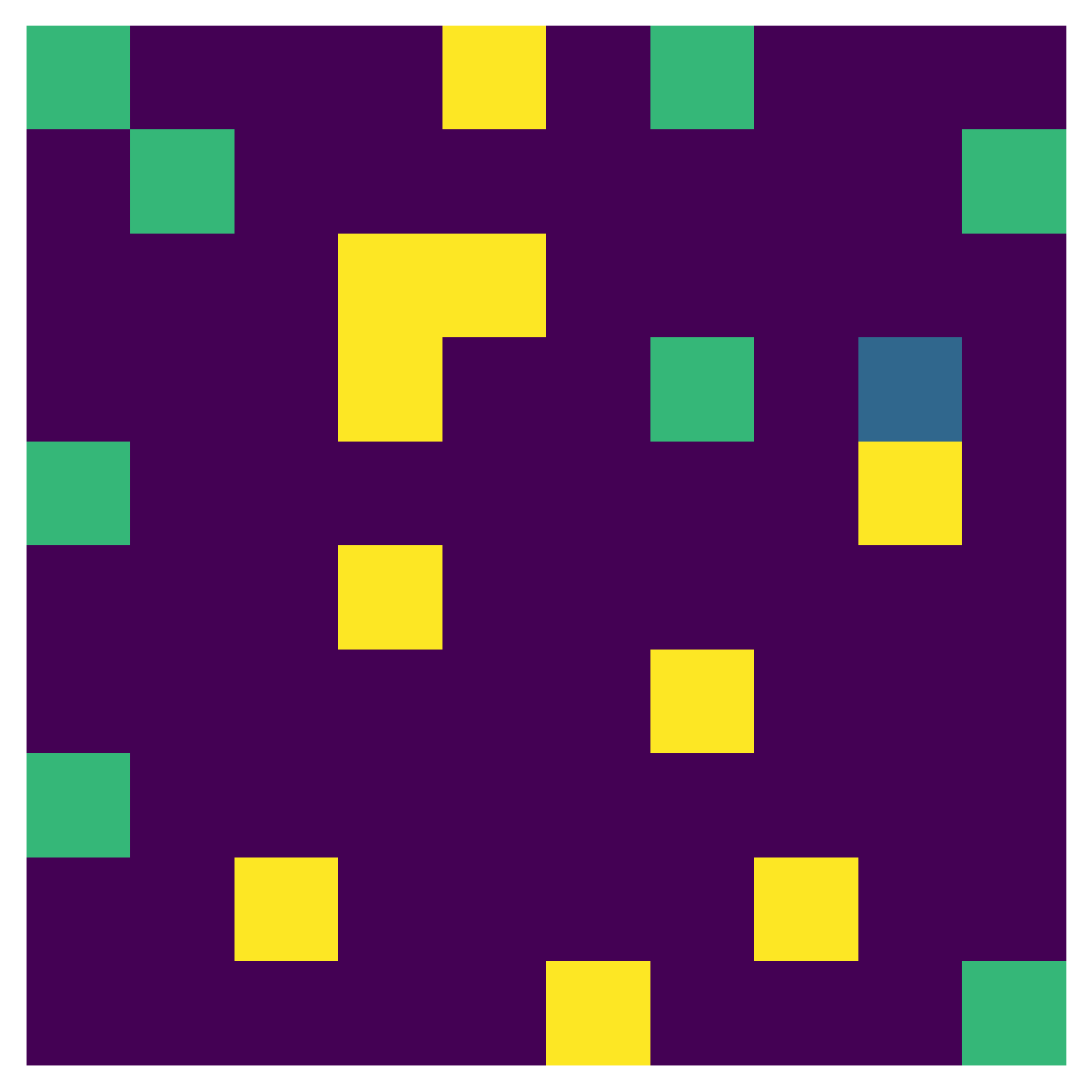}
\includegraphics[width=.95\linewidth]{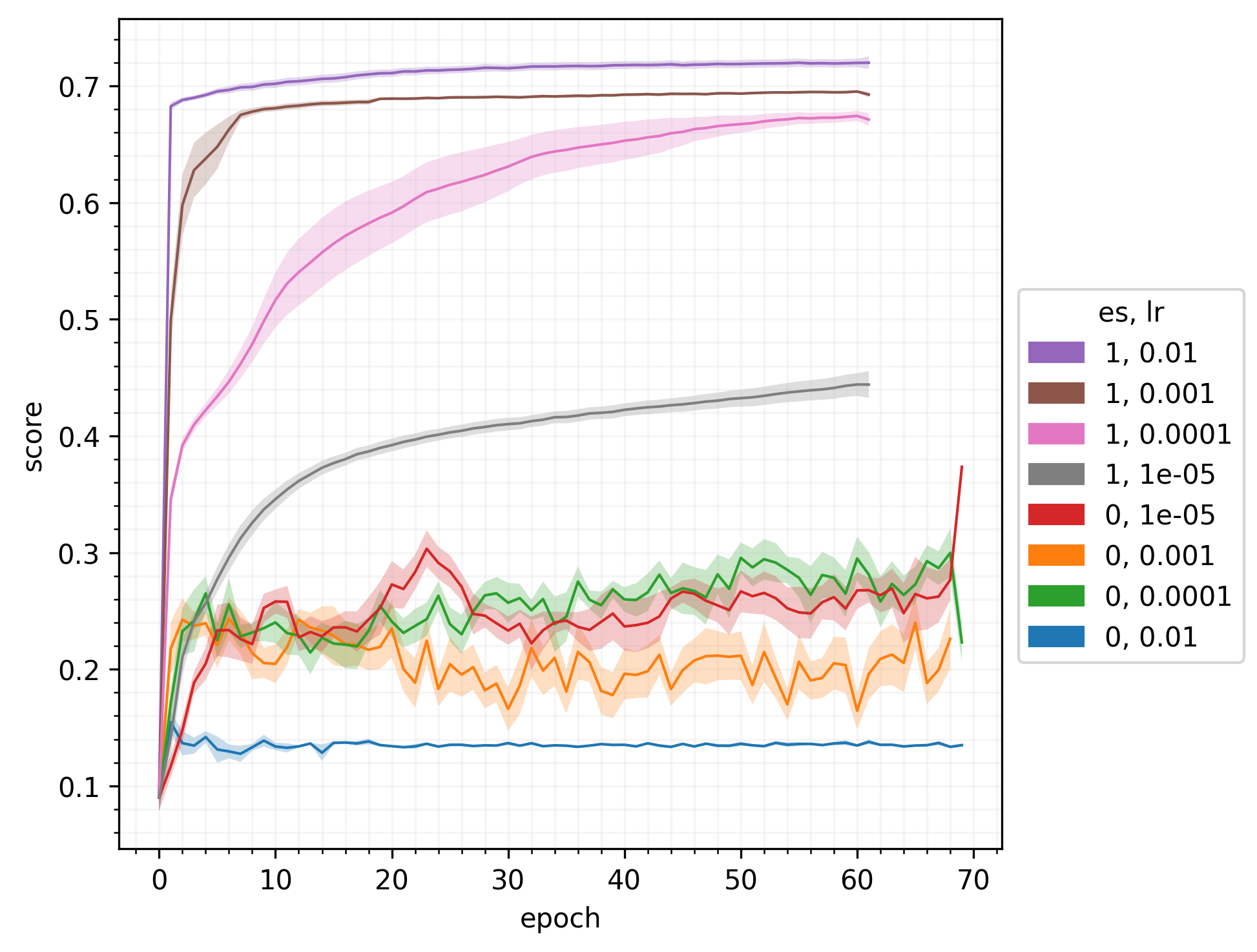}
\includegraphics[width=.95\linewidth]{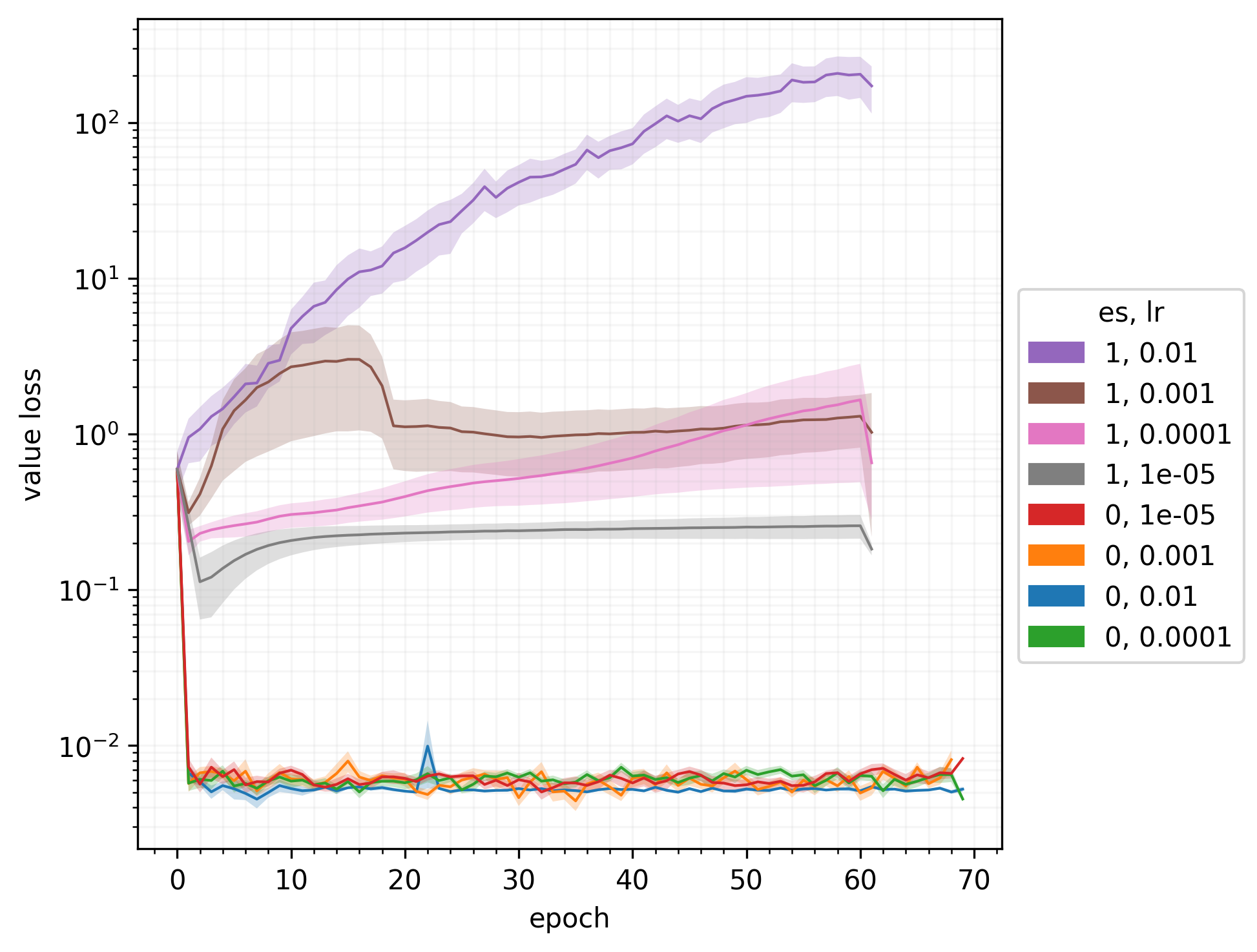}
\includegraphics[width=.95\linewidth]{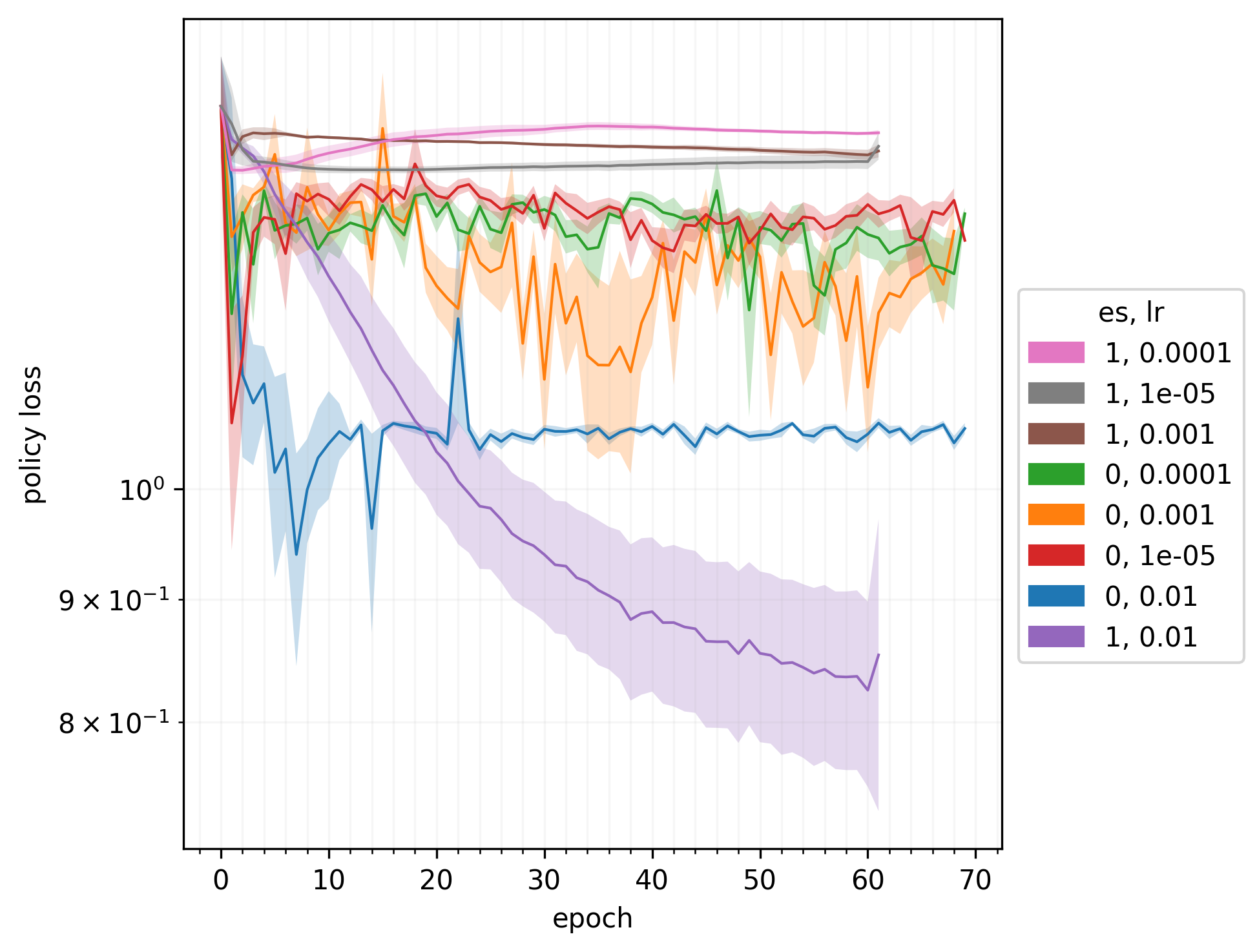}
\caption{Navigation state and metrics.}
\label{fig:navigation}
\end{figure}

An example state is shown in Figure~\ref{fig:navigation}.
The blue square is the agent.
Green squares are unreached targets.
Yellow squares are reached targets.
Experimental results are shown in Figure \ref{fig:navigation}.
AlphaZeroES outperforms AlphaZero in terms of episode score.
Unlike AlphaZero, it does not seem to minimize the value and policy losses by a noticeable amount.

\paragraph{Sokoban}

Sokoban is a puzzle in which an agent pushes boxes around a warehouse to get them to storage locations.
It is played on a grid of tiles.
Each tile may be a floor or a wall, and may contain a box or the agent.
Some floor tiles are marked as storage locations.
The agent can move horizontally or vertically onto empty tiles.
The agent can also move a box by walking up to it and push it to the tile beyond, if the latter is empty.
Boxes cannot be pulled, and they cannot be pushed to squares with walls or other boxes.
The number of boxes equals the number of storage locations.
The puzzle is solved when all boxes are placed at storage locations.
Planning ahead is crucial, since an agent can easily get stuck if it makes the wrong move.

Sokoban has been studied in the field of computational complexity and shown to be PSPACE-complete~\citep{sokoban_pspace_complete}.
It has received significant interest in artificial intelligence research because of its relevance to automated planning (\emph{e.g.}, for autonomous robots), and is used as a benchmark.
Sokoban's large branching factor and search tree depth contribute to its difficulty.
Skilled human players rely mostly on heuristics and can quickly discard several futile or redundant lines of play by recognizing patterns and subgoals, narrowing down the search significantly.
Various automatic solvers have been developed in the literature~\citep{junghanns1997sokoban,junghanns2001sokoban,froleyks2016using,shoham2020fess}, many of which rely on heuristics, but more complex Sokoban levels remain a challenge.

Our environment is as follows.
We use the unfiltered Boxoban training set~\citep{boxobanlevels}, which contains 900,000 levels of size \(10 \times 10\) each.
At the beginning of each episode, we sample a level from this dataset uniformly at random.
As a form of data augmentation, we sample one of the eight symmetries of the square uniformly at random (a horizontal flip, vertical flip, and/or 90-degree rotation) and apply it to the level.
In each timestep, the agent has four actions available to it, for motion in each of the four cardinal directions.
The level ends after a specified number of timesteps.
(We use 50 timesteps.)
The return at the end of an episode is the number of goals that are covered with boxes.
Thus the agent is incentivized to cover all of the goals.
The prediction network observes a set of vectors, one for each tile in the level, where each vector contains the 2 coordinates of the tile, 4 boolean flags indicating whether the tile contains a wall, goal, box, and/or agent, and the number of episode timesteps remaining.

\begin{figure}
\centering
\includegraphics[width=.6\linewidth]{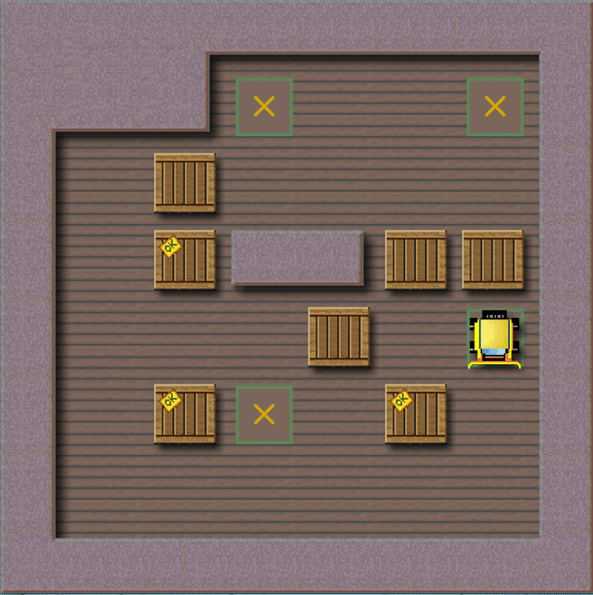}
\includegraphics[width=.95\linewidth]{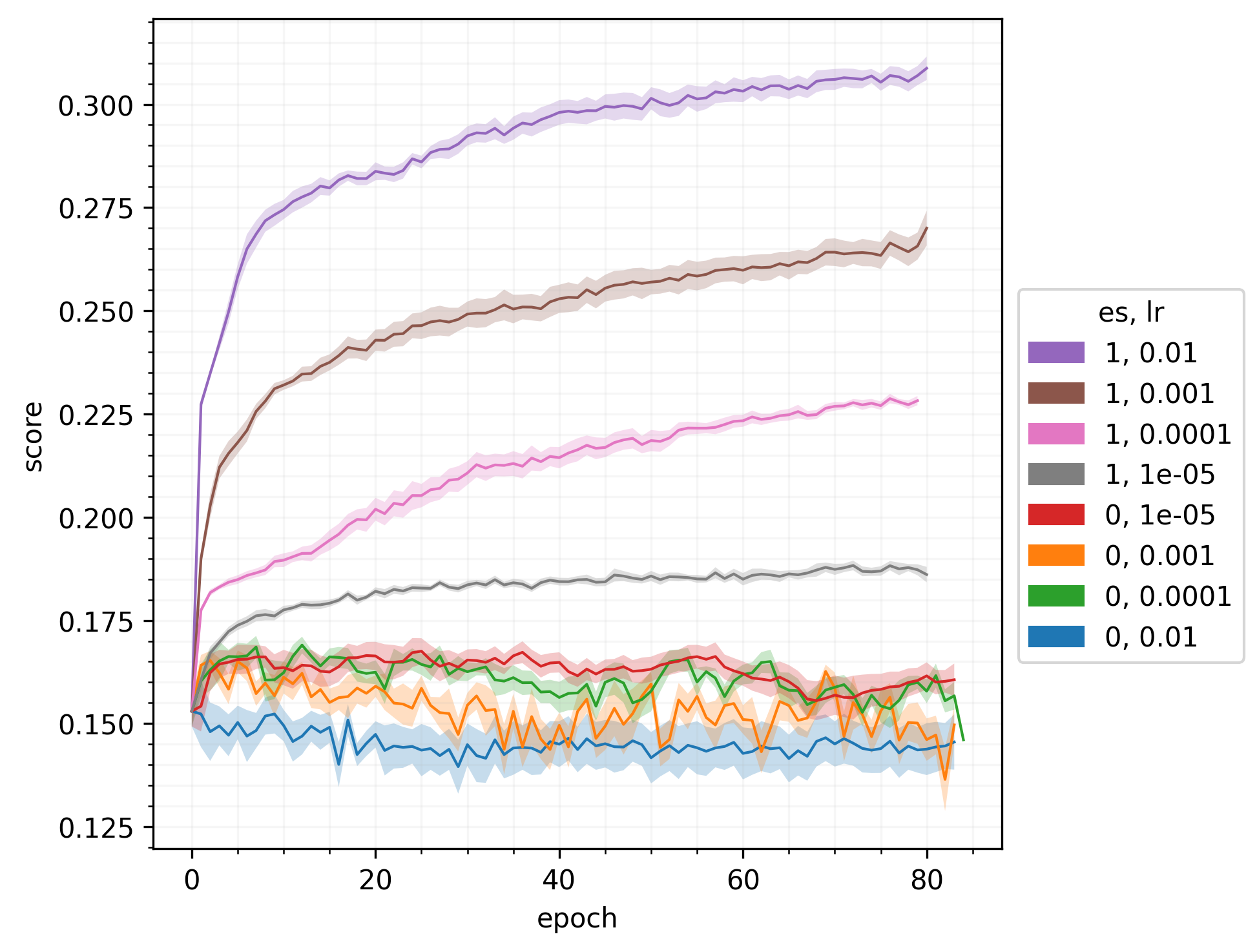}
\includegraphics[width=.95\linewidth]{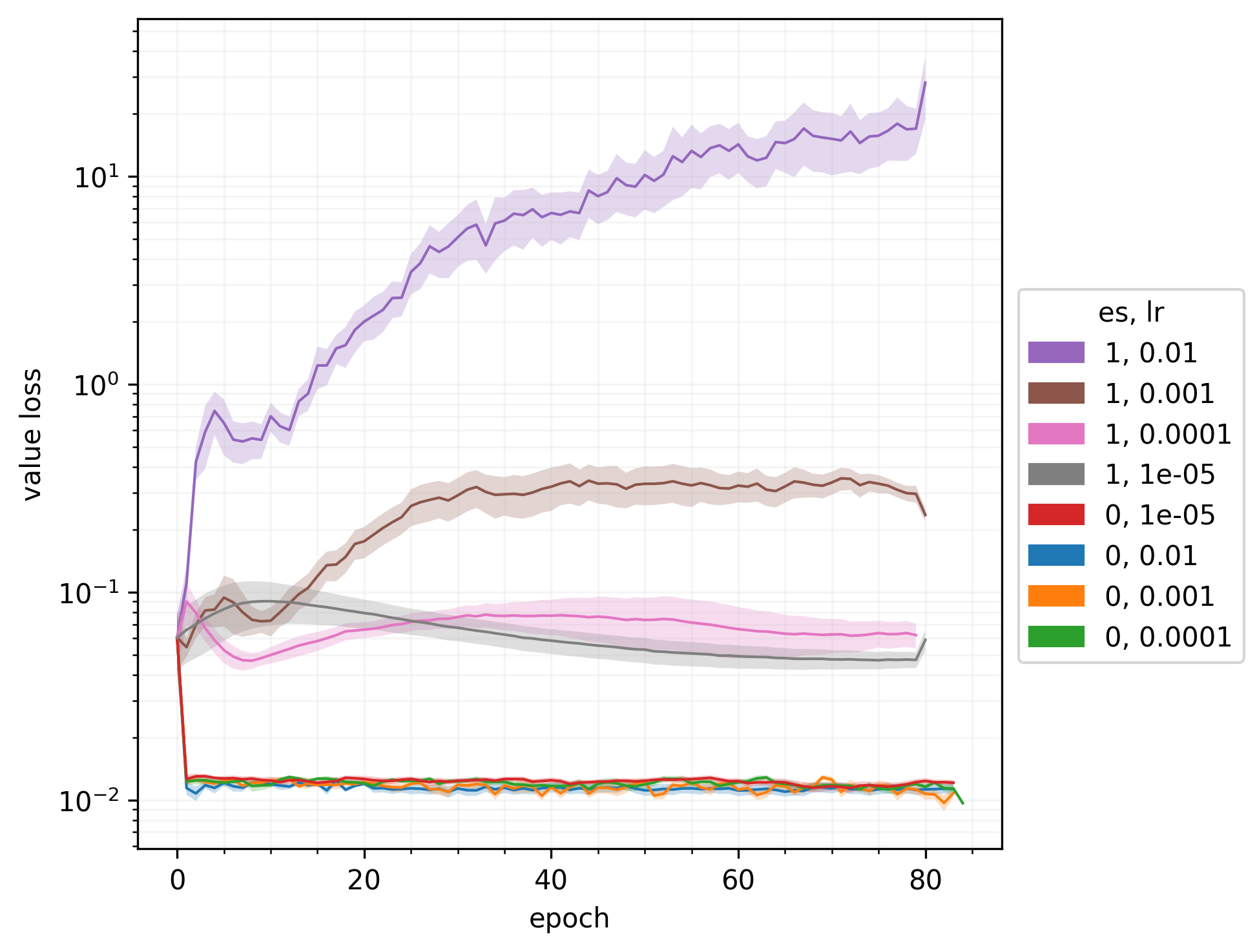}
\includegraphics[width=.95\linewidth]{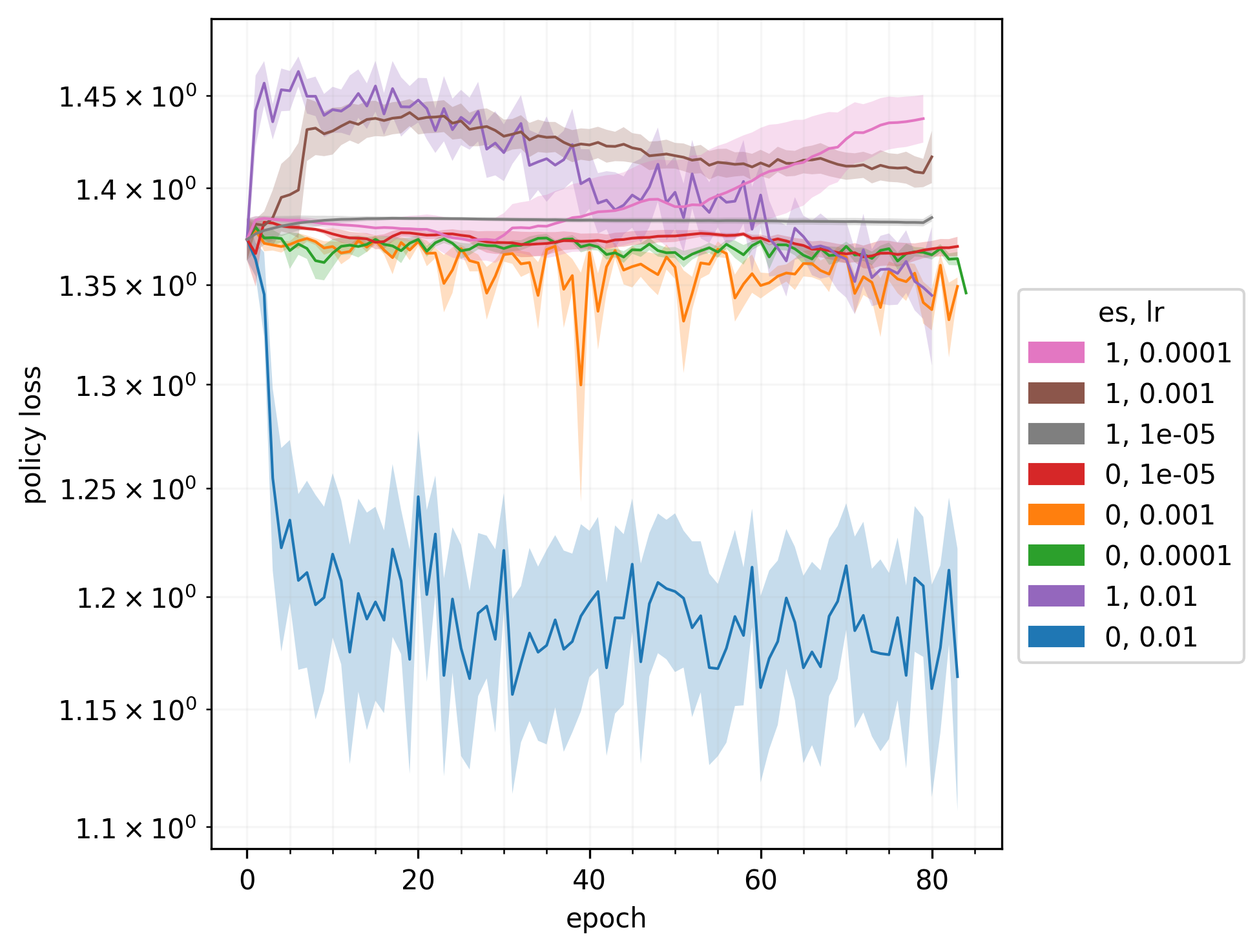}
\caption{Sokoban state and metrics.}
\label{fig:sokoban}
\end{figure}

An example state is shown in Figure~\ref{fig:sokoban}.
This image was rendered by the JSoko software \citep{jsoko}, an open-source Sokoban implementation.
The yellow vehicle is the agent, who must push the brown boxes into the goal squares marked with Xs.
(Boxes tagged ``OK'' are on top of goal squares.)
Experimental results are shown in Figure \ref{fig:sokoban}.
AlphaZeroES outperforms AlphaZero in terms of episode score.
Unlike AlphaZero, it does not seem to minimize the value and policy losses by a noticeable amount.

\paragraph{Traveling salesman problem}

The \emph{traveling salesman problem (TSP)} is a classic CO problem.
Given a set of cities and their pairwise distances, the goal is to find a shortest route that visits each city once and returns to the starting city.
This problem has important applications in operations research, including logistics, computer wiring, vehicle routing, and various other planning problems~\citep{matai2010traveling}.
TSP is known to be NP-hard~\citep{karp1972reducibility}, even in the Euclidean setting \citep{papadimitriou1977euclidean}.
Various approximation algorithms and heuristics~\citep{nilsson2003heuristics} have been developed for it.

Our environment is as follows.
We seek to learn to solve TSP in general, not just one particular instance of it.
Thus, on every episode, a new problem instance is generated by sampling a matrix \(\mathbf{X} \sim \operatorname{Uniform}([0, 1]^{n \times 2})\), representing a sequence of \(n \in \mathbb{N}\) cities.
In our experiments, we use \(n = 20\).
At timestep \(t \in [n]\), the agent chooses a city \(a_t \in [n]\) that has not been visited yet.
At the end of the episode, the length of the tour through this sequence of cities (including the segment from the final city to the initial one) is computed, and treated as the \emph{negative} score.
Thus the agent is incentivized to find the shortest tour through all the cities.
Formally, the final score is \(-\sum_{t \leq n} d(\mathbf{X}_{a_t}, \mathbf{X}_{a_{t+1} \bmod n})\), where \(d\) is the Euclidean metric.
The prediction network observes a set of vectors, one for each city, where each vector contains the coordinates of the city and 3 boolean 0-1 flags indicating whether it has already been visited, whether it is the initial city, and whether it is the current city.

\begin{figure}
\centering
\includegraphics[width=.6\linewidth]{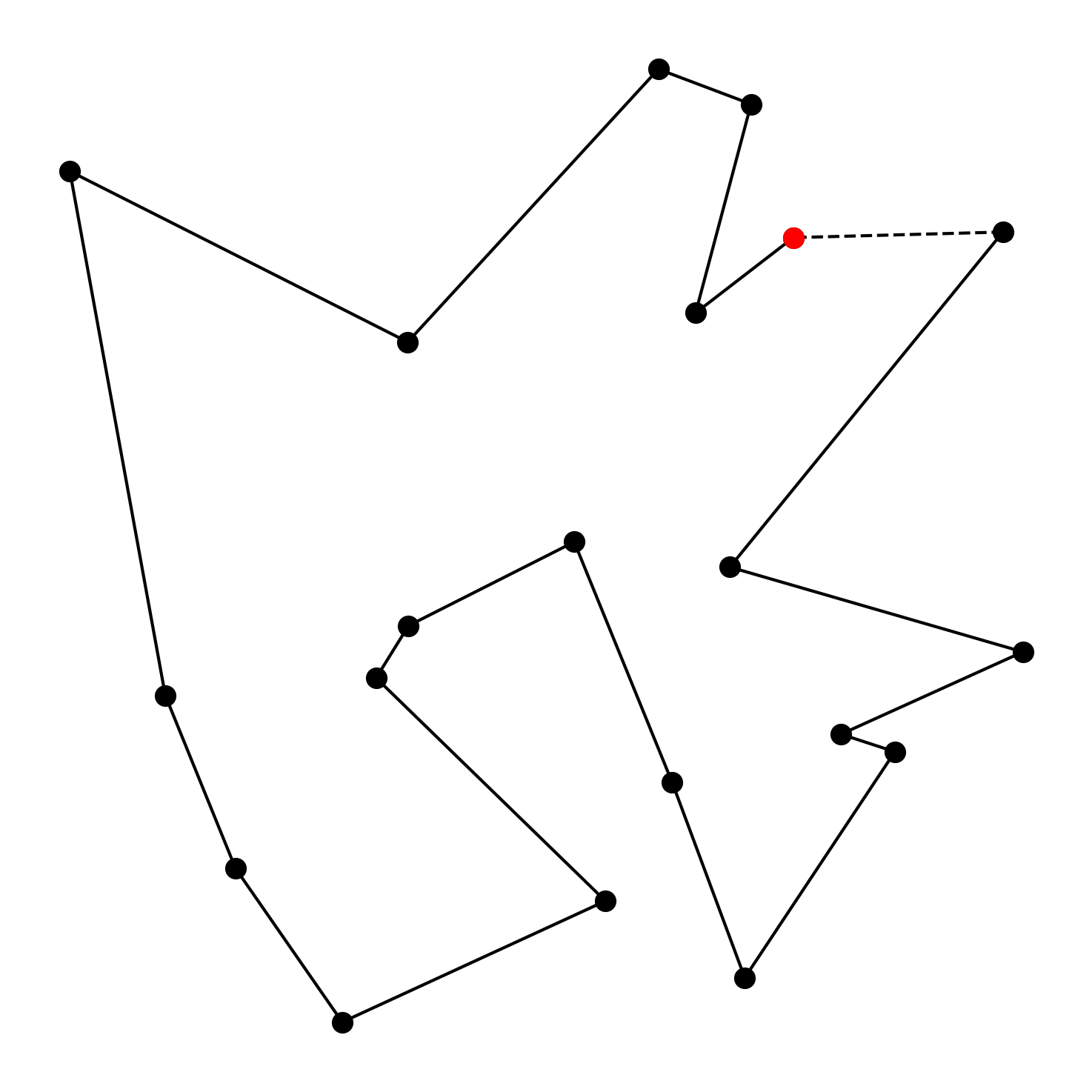}
\includegraphics[width=.95\linewidth]{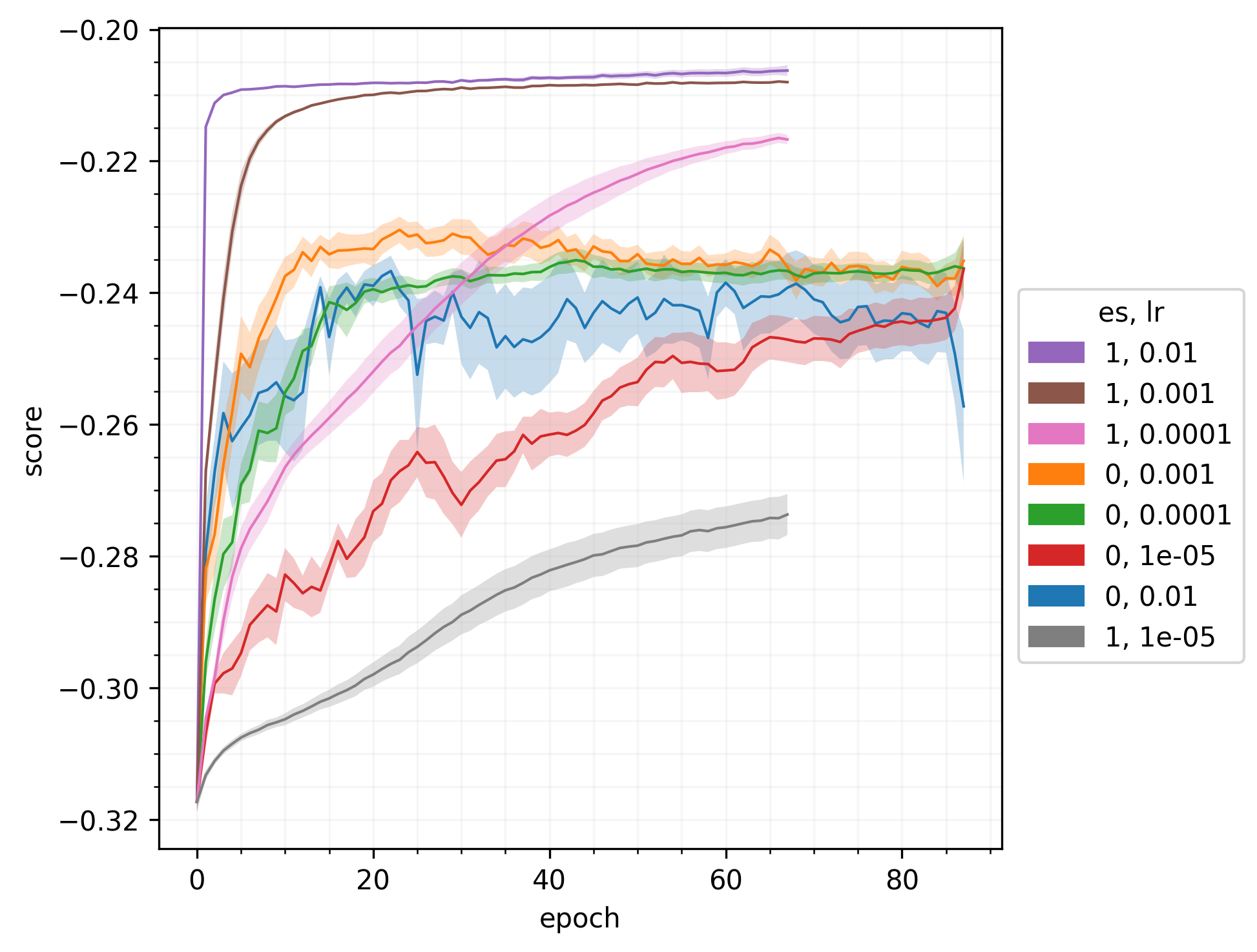}
\includegraphics[width=.95\linewidth]{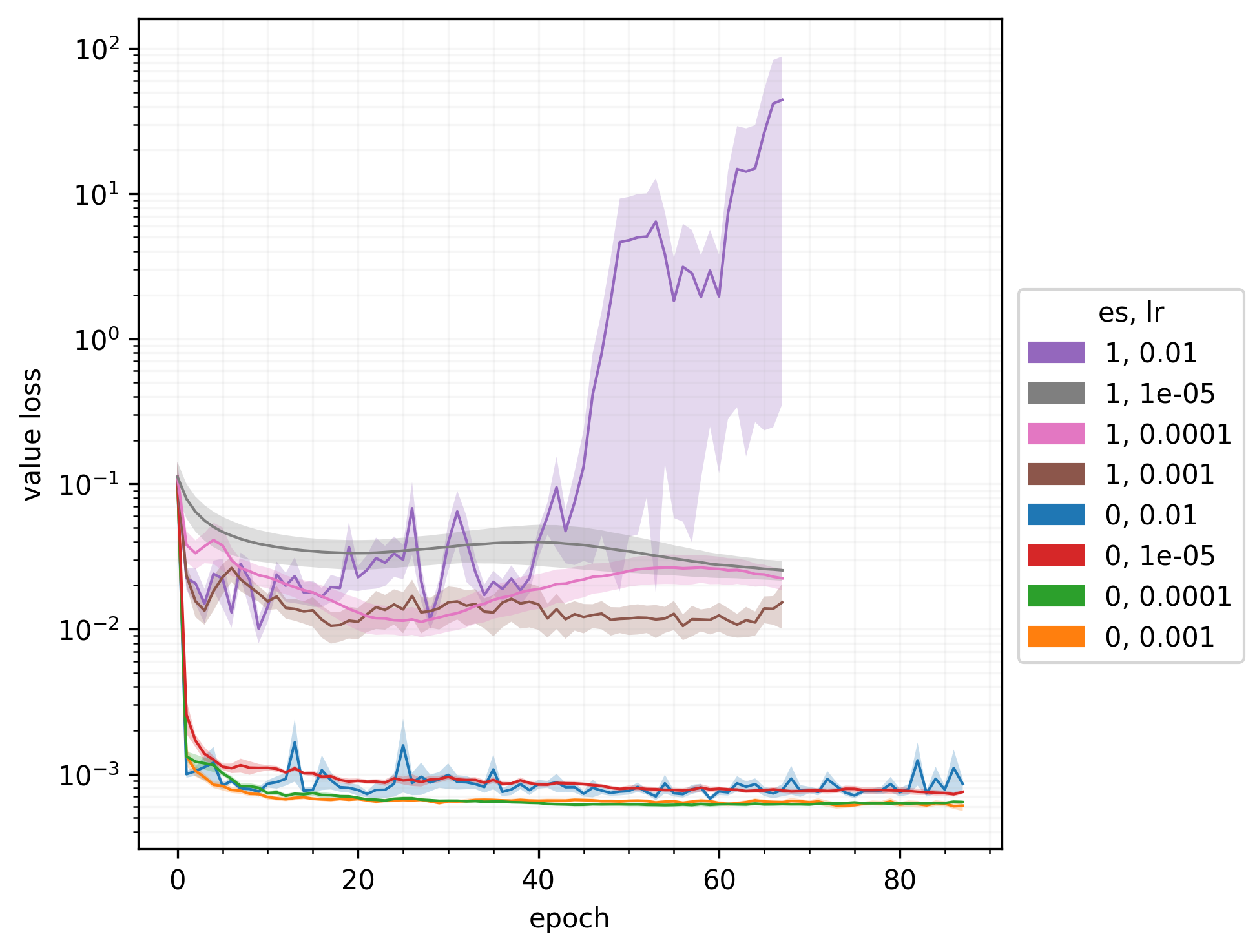}
\includegraphics[width=.95\linewidth]{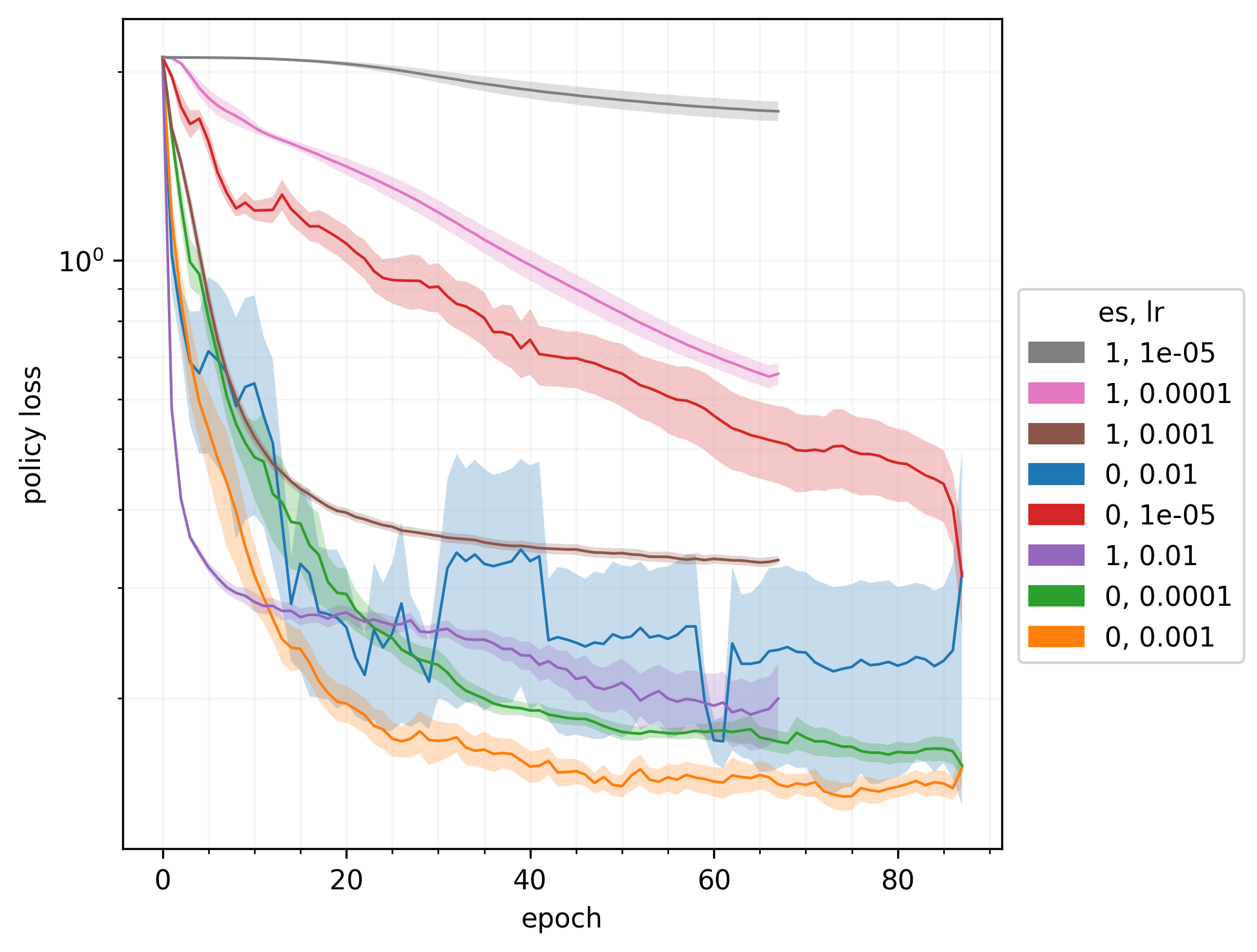}
\caption{TSP state and metrics.}
\label{fig:tsp}
\end{figure}

An example state is shown in Figure~\ref{fig:tsp}.
Dots are cities.
The red dot is the initial city.
The lines connecting the dots constitute the constructed path.
The dotted line is the last leg from the final city back to the initial city.
Experimental results are shown in Figure \ref{fig:tsp}.
AlphaZeroES outperforms AlphaZero in terms of episode score.
Interestingly, as a side effect, it minimizes the policy loss about as much as AlphaZero does.
It also minimizes the value loss (except at the highest learning rate), though to a lesser extent than AlphaZero.

\paragraph{Vertex k-center problem}

The \emph{vertex k-center problem (VKCP)} is a classic CO problem that has applications in facility location and clustering.
The problem is as follows.
Given \(n\) points in \(\mathbb{R}^d\), select a subset \(\mathcal{S}\) of \(k\) points that maximizes the distance from any point in the original set to its nearest point in \(\mathcal{S}\).
The \(n\) points can be interpreted as possible locations in which to build facilities (\emph{e.g.}, fire stations, police stations, supply depots, \emph{etc.}), where \(\mathcal{S}\) is the set of locations in which such facilities are built, and the goal is to minimize the maximum distance from any location to its nearest facility.
(There is also a variant of the problem that seeks to minimize the \emph{mean} distance.)
This problem was first proposed by \citet{hakimi1964optimum}.
It is an NP-hard problem, and various approximation algorithms have been proposed for it \citep{kariv1979algorithmic,gonzalez1985clustering,dyer1985simple,hochbaum1985best,shmoys1994computing}.
A survey and evaluation of approximation algorithms can be found in \citet{garcia2019approximation}.

For our experiments, we sample \(n = 40\) locations uniformly at random from the unit square and let \(k = 20\).
At any timestep \(t\), the agent can select a location \(a_t \in [n]\) that has not been selected yet to add a facility at that location.
The final score is \(-\max_{i \in [n]} \min_{j \in \mathcal{S}} d(\mathbf{x}_i, \mathbf{x}_j)\), where \(\mathbf{x}_i \in [0, 1]^2\) is the position of point \(i \in [n]\) and \(d\) is the Euclidean metric.
The prediction network observes a set of vectors, one for each point, where each vector contains the coordinates of the point and a boolean 0-1 flag indicating whether it has been included in the set.

\begin{figure}
\centering
\includegraphics[width=.6\linewidth]{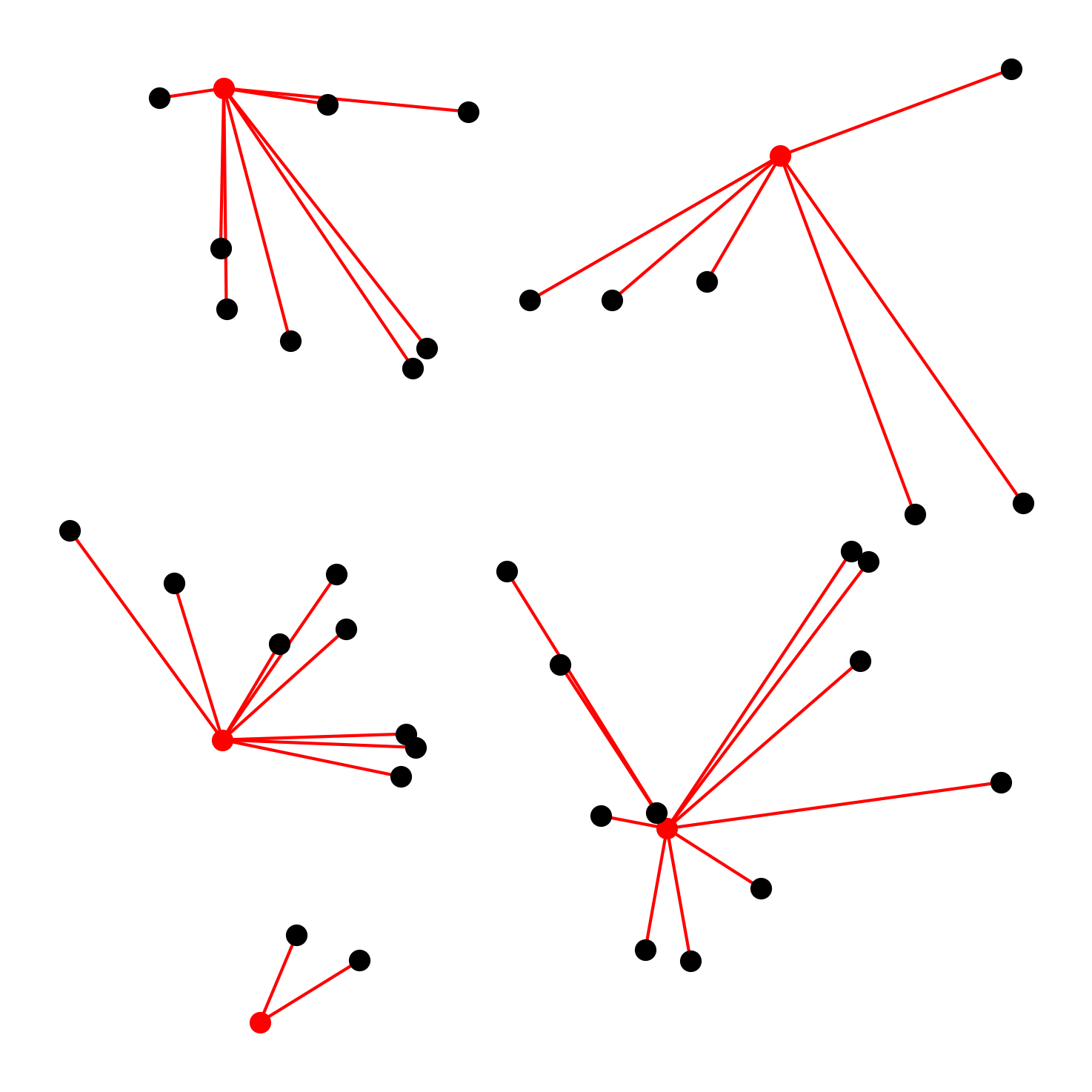}
\includegraphics[width=.95\linewidth]{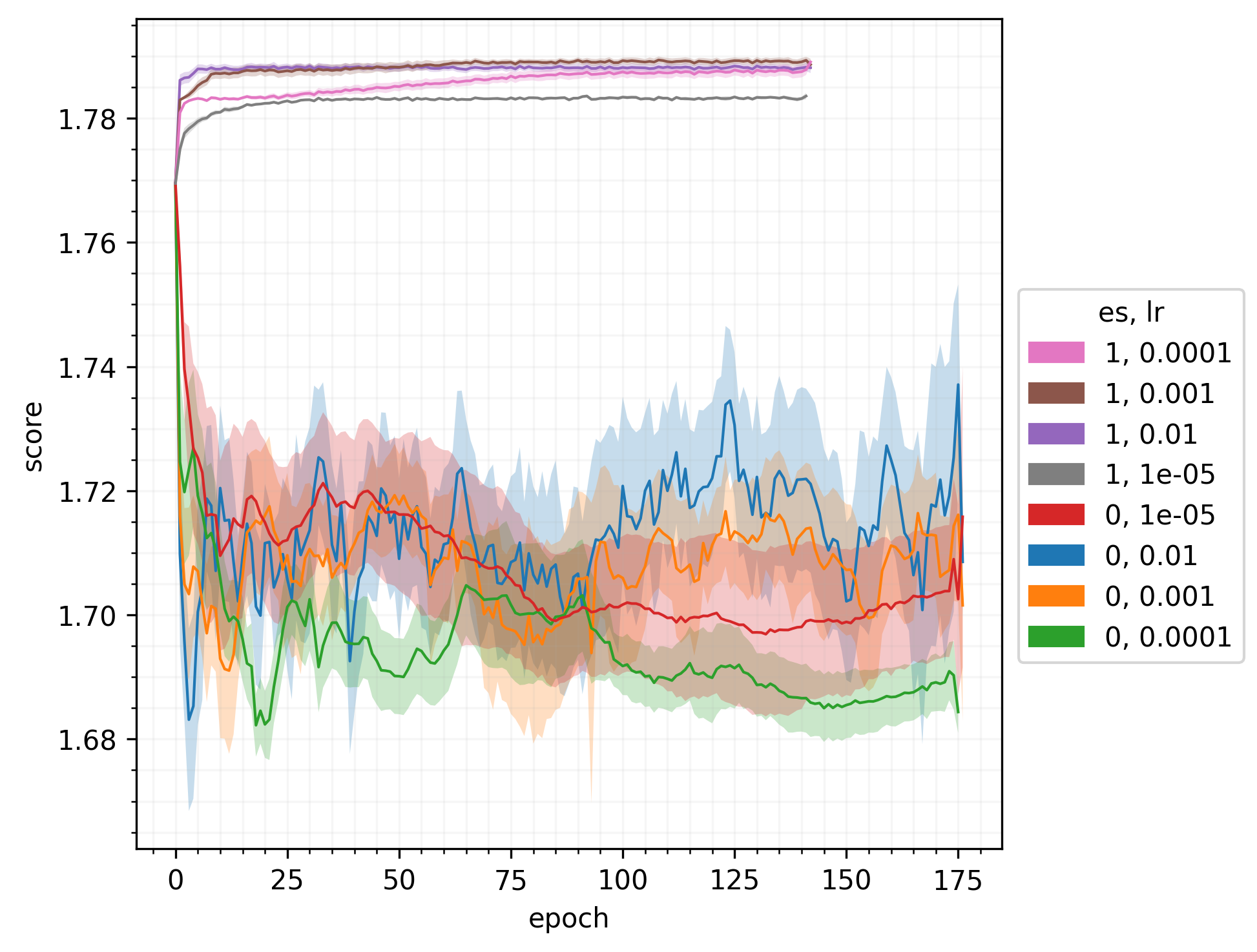}
\includegraphics[width=.95\linewidth]{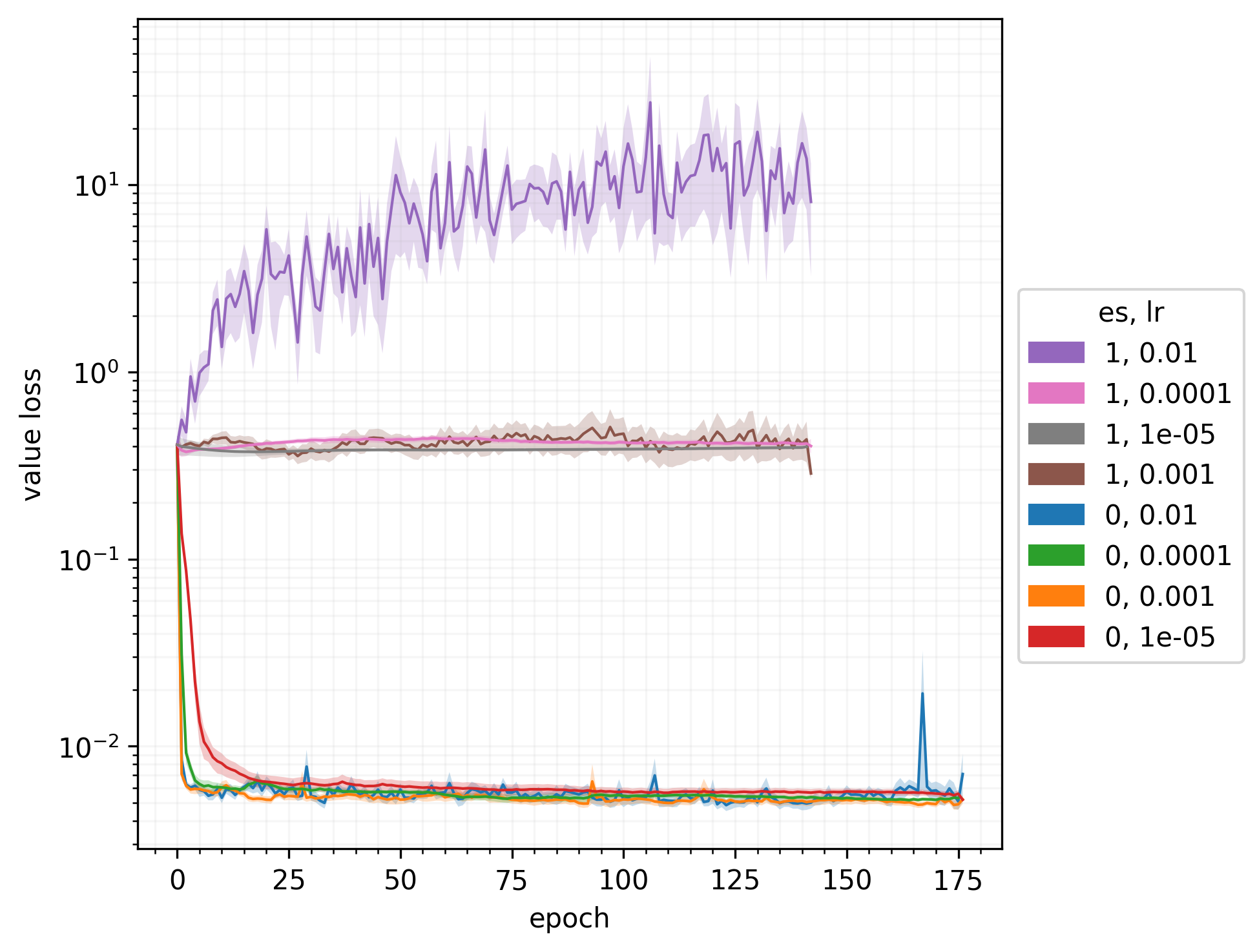}
\includegraphics[width=.95\linewidth]{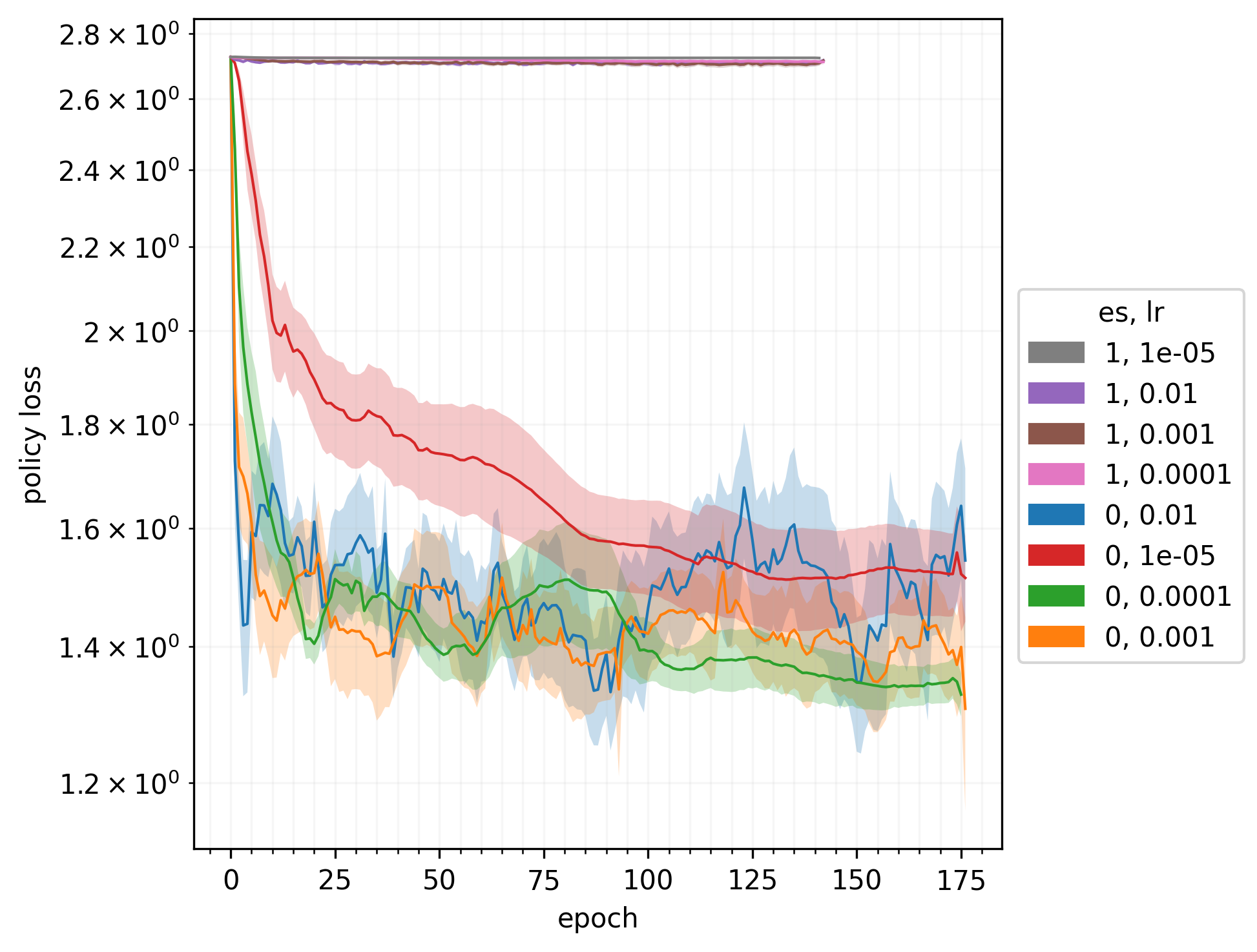}
\caption{VKCP state and metrics.}
\label{fig:vertex_center}
\end{figure}

An example state is shown in Figure \ref{fig:vertex_center}.
Black dots are locations, red dots are facilities placed so far, and red lines connect locations to their nearest facility.
Experimental results are shown in Figure \ref{fig:vertex_center}.
AlphaZeroES outperforms AlphaZero in terms of episode score.
In this environment, AlphaZeroES hardly minimizes the value and policy losses as a side effect.

\paragraph{Maximum diversity problem}

The \emph{maximum diversity problem (MDP)} is as follows.
Given \(n\) points in \(\mathbb{R}^d\), select a subset \(\mathcal{S}\) of \(k\) points that maximizes the minimum distance between distinct points.
(There is also a variant of the problem that seeks to maximize the \emph{mean} distance between distinct points.)
This problem is strongly NP-hard, as can be shown via reduction from the clique problem \citep{kuo1993analyzing,ghosh1996computational}.
Various heuristics have been proposed for it \citep{glover1998heuristic,katayama2005evolutionary,silva2007new,duarte2007tabu,marti2010branch,lozano2011iterated,wu2013hybrid,marti2013heuristics}.
This problem has applications in ecology, medical treatment, genetic engineering, capital investment, pollution control, system reliability, telecommunication services, molecular structure design, transportation system control, emergency service centers, and energy options, as catalogued in \citet[Table 1]{glover1998heuristic}.

For our experiments, we sample \(n = 40\) locations uniformly at random from the unit square and let \(k = 20\).
At any timestep \(t\), the agent can select a point \(a_t \in [n]\) that has not been selected yet to add to the set \(\mathcal{S}\).
The final score is \(\min_{i, j \in \mathcal{S}, i \neq j} d(\mathbf{x}_i, \mathbf{x}_j)\), where \(\mathbf{x}_i \in [0, 1]^2\) is the position of point \(i\) and \(d\) is the Euclidean metric.
The prediction network observes a set of vectors, one for each point, where each vector contains the coordinates of the point and a boolean 0-1 flag indicating whether it has been included in the set.

\begin{figure}
\centering
\includegraphics[width=.6\linewidth]{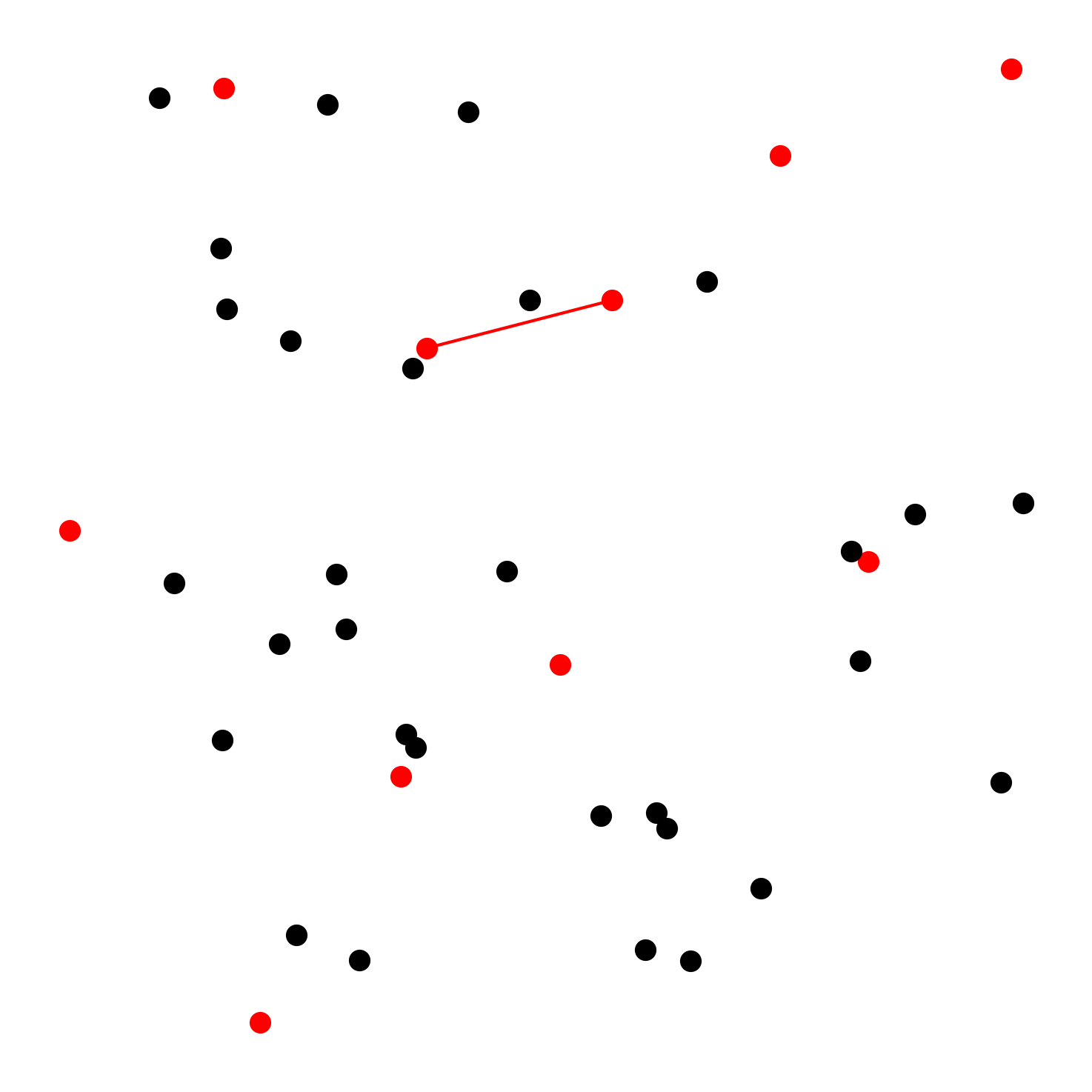}
\includegraphics[width=.95\linewidth]{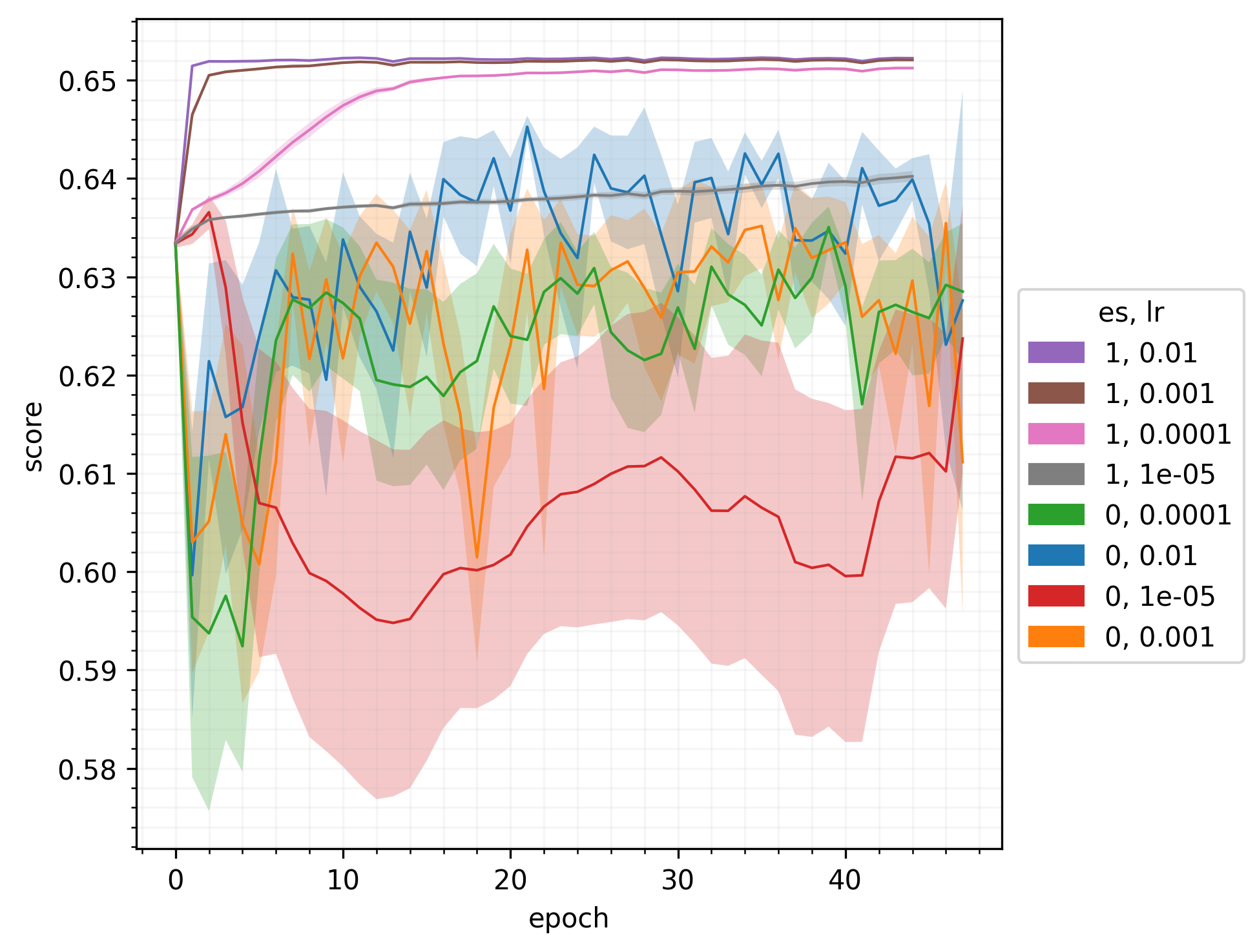}
\includegraphics[width=.95\linewidth]{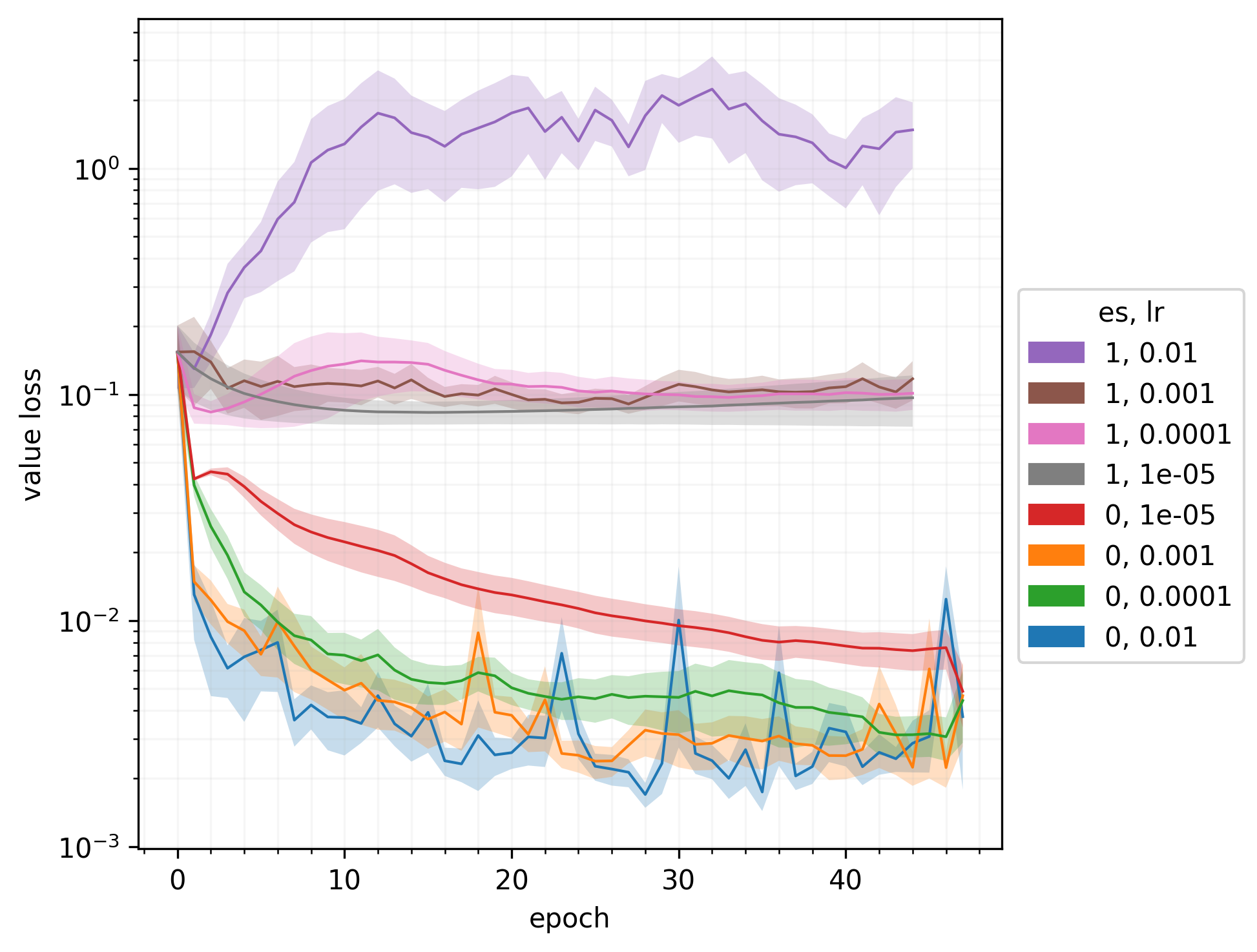}
\includegraphics[width=.95\linewidth]{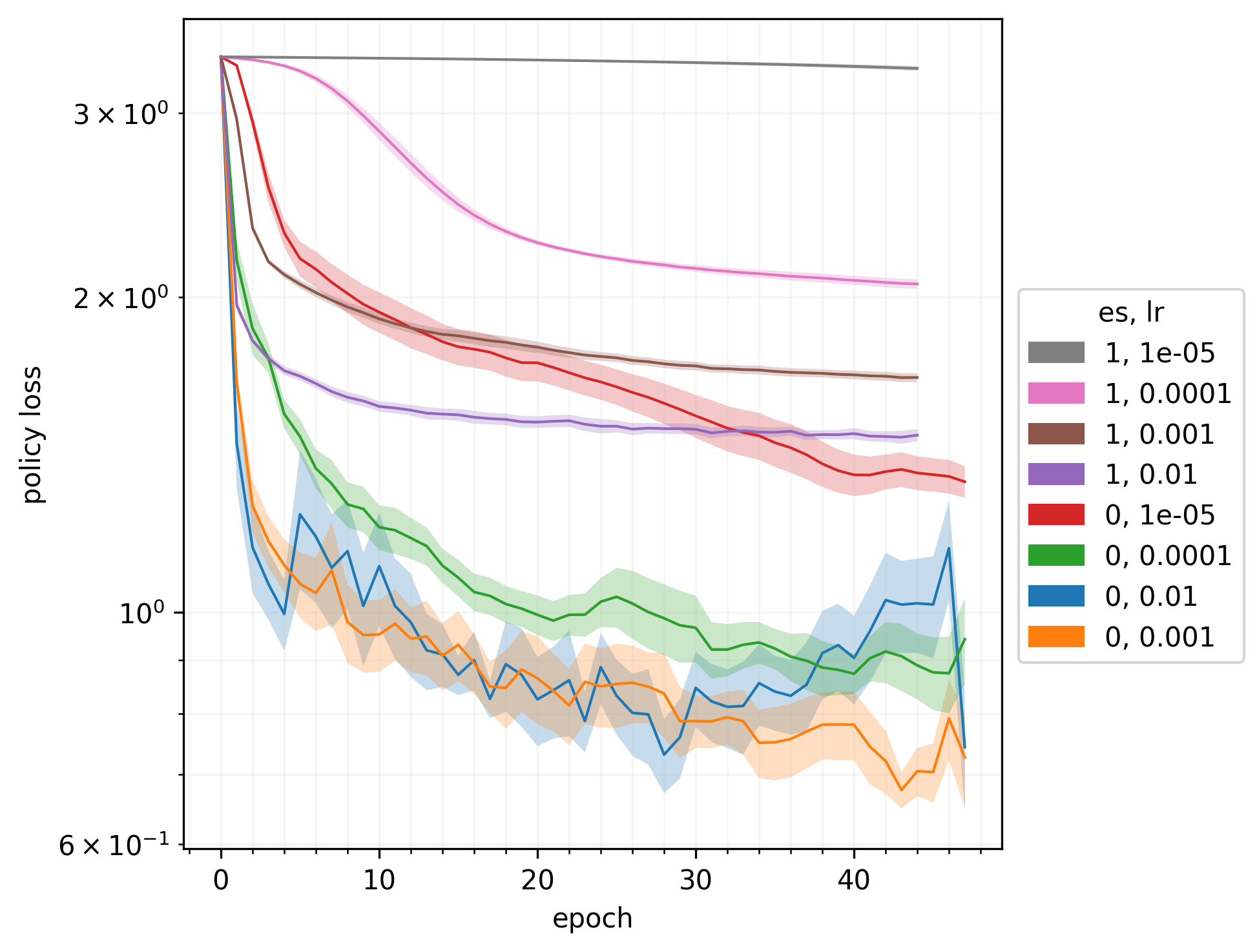}
\caption{MDP state and metrics.}
\label{fig:max_diversity}
\end{figure}

An example state is shown in Figure~\ref{fig:max_diversity}.
Black dots are the points, red dots are the points selected so far, and the red line connects the closest pair of distinct points in the set selected so far.
Experimental results are shown in Figure \ref{fig:max_diversity}.
AlphaZeroES outperforms AlphaZero in terms of episode score.
As a side effect, it minimizes the policy loss about as much as AlphaZero does.
However, unlike AlphaZero, it does not seem to minimize the value loss.

\section{Conclusion}
\label{sec:conclusion}

In this paper, we set out to study whether AlphaZero and its newest variants can be improved by maximizing the episode score directly instead of minimizing the standard planning loss.
Since MCTS is not differentiable, we maximize the episode score by using evolution strategies.
We tested our approach on multiple CO problems and motion planning problems.
Our experimental results indicate that our approach outperforms planning loss minimization.
Our work opens up new possibilities for tackling environments where planning is important.
It does this by allowing agents to learn to leverage internal nondifferentiable planning algorithms, such as MCTS, \emph{in a purely blackbox way}.
That is, instead of training the agent's parameters to minimize some indirect proxy objective, such as a planning loss, we can now maximize the desired objective \emph{directly}.

\section{Acknowledgments}

This material is based on work supported by the Vannevar Bush Faculty Fellowship ONR N00014-23-1-2876, National Science Foundation grants RI-2312342 and RI-1901403, ARO award W911NF2210266, and NIH award A240108S001.

\bibliographystyle{plainnat}
\bibliography{dairefs,references}
\appendix
\section{Additional related research}

\emph{Value Iteration Network (VIN)}~\citep{vin} is a fully differentiable network with a planning module embedded within. It can learn to plan and predict outcomes that involve planning-based reasoning, such as policies for reinforcement learning. It uses a differentiable approximation of the value-iteration algorithm, which can be represented as a convolutional network, and is trained end-to-end using standard backpropagation.

Predictron~\citep{predictron} consists of a fully abstract model, represented by a Markov reward process, that can be rolled forward multiple ``imagined'' planning steps.
Each forward pass accumulates internal rewards and values over multiple planning depths.
The model is trained end-to-end so as to make these accumulated values accurately approximate the true value function.

\emph{Value Prediction Network (VPN)}~\citep{vpn} integrates model-free and model-based RL methods into a single network.
In contrast to previous model-based methods, it learns a dynamics model with abstract states that is trained to make action-conditional predictions of future returns rather than future observations.
VIN performs value iteration over the entire state space, which requires that 1) the state space is small and representable as a vector with each dimension corresponding to a separate state and 2) the states have a topology with local transition dynamics (such as a 2D grid).
VPN does not have these limitations.
VPN is trained to make its predicted values, rewards, and discounts match up with those of the real environment~\citep[\S 3.3]{vpn}.

\emph{Imagination-Augmented Agent (I2A)}~\citep{racaniere2017imagination} augments a model-free agent with imagination by using environment models to simulate imagined trajectories, which are provided as additional context to a policy network.
An environment model is any recurrent architecture which can be trained in an unsupervised fashion from agent trajectories.
Given a past state and current action, the environment model predicts the next state and observation.
The imagined trajectory is initialized with the current observation and rolled out multiple time steps into the future by feeding simulated observations.

MCTSnet~\citep{mcts_nets} incorporates simulation-based search inside a neural network, by expanding, evaluating and backing-up a vector embedding.
The parameters of the network are trained end-to-end using gradient-based optimisation.
When applied to small searches in the well-known planning problem Sokoban, it outperforme dprior MCTS baselines.

TreeQN~\citep{treeqn} is an end-to-end differentiable architecture that substitutes value function networks in discrete-action domains.
Instead of directly estimating the state-action value from the current encoded state, as in \emph{Deep Q-Networks (DQN)}~\citep{dqn}, it uses a learned dynamics model to perform planning up to some fixed-depth.
The result is a recursive, tree-structured network between the encoded state and the predicted state-action values at the leafs.
The authors also propose ATreeC, an actor-critic variant that augments TreeQN with a softmax layer to form a stochastic policy network.
Unlike MCTS-based methods, the shape of the planning tree is fixed, and the agent cannot ``focus'' on more promising subtrees to expand during planning.

\citet{yang2020continuous} proposed Continuous MuZero, an extension of MuZero to continuous actions, and showed that it outperforms the \emph{soft actor-critic (SAC)} algorithm.
\citet{hubert2021learning} proposed Sampled MuZero, an extension of the MuZero algorithm that is able to learn in domains with arbitrarily complex action spaces (including ones that are continuous and high-dimensional) by planning over sampled actions.

Stochastic MuZero~\citep{stochastic_muzero} extended MuZero to environments that are inherently stochastic, partially observed, or so large and complex that they appear stochastic to a finite agent.
It learns a stochastic model incorporating after-states following an action, and uses this model to perform a stochastic tree search.
It matches the performance of MuZero in Go while matching or exceeding the state of the art in a set of canonical single and multiagent environments, including 2048 and backgammon.

\subsubsection*{Machine learning for tuning integer programming and combinatorial optimization techniques}

Another, different, form of learning in search techniques is tuning \emph{integer programming (IP)} and \emph{combinatorial optimization (CO)} \citep{schrijver2003combinatorial} techniques.
The idea of automated algorithm tuning goes back at least to \citet{rice1976algorithm}.
It has been applied in industrial practice at least since 2001, when \citet{sandholm12013very} started using machine learning to learn IP algorithm configurations (related to branching, cutting plane generation, \emph{etc}.) and IP formulations based on problem instance features, in the context of combinatorial auction winner determination in large-scale sourcing auctions.
In 2007, the leading commercial general-purpose IP solvers started shipping with such automated configuration tools.

IP solvers typically use a tree search algorithm called branch-and-cut.
However, such solvers typically come with a variety of tunable parameters that are challenging to tune by hand. 
Research has demonstrated the power of using a data-driven approach to automatically optimize these parameters. 

Similarly, real-world applications that can be formulated as CO problems often have recurring patterns or structure that can be exploited by heuristics.
The design of good heuristics or approximation algorithms for NP-hard CO problems often requires significant specialized knowledge and trial-and-error, which can be a challenging and tedious process.

The rest of this section reviews some of the newer work on automated algorithm configuration in IP and CO.

\citet{khalil2017learning} sought to automate the CO tuning process using a combination of reinforcement learning and graph embedding.
They applied their framework to a diverse range of optimization problems over graphs, learning effective algorithms for the Minimum Vertex Cover, Maximum Cut and Traveling Salesman problems.

\citet{bengio2021machine} surveyed recent attempts from the machine learning and operations research communities to leverage machine learning to solve IP and CO problems.
According to the authors,
``Given the hard nature of these problems, state-of-the-art algorithms rely on handcrafted heuristics for making decisions that are otherwise too expensive to compute or mathematically not well defined.
Thus, machine learning looks like a natural candidate to make such decisions in a more principled and optimized way.''
They cite \citet{larsen2018predicting}, who train a neural network to predict the solution of a stochastic load planning problem for which a deterministic mixed integer linear programming formulation exists.
The authors state that
``The nature of the application requires to output solutions in real time, which is not possible either for the stochastic version of the load planning problem or its deterministic variant when using state-of-the-art MILP solvers.
Then, ML turns out to be suitable for obtaining accurate solutions with short computing times because some of the complexity is addressed offline, \textit{i.e.}, in the learning phase, and the run-time (inference) phase is extremely quick.''

Another survey of reinforcement learning for CO can be found in \citet{mazyavkina2021reinforcement}.
According to the authors,
``Many traditional algorithms for solving combinatorial optimization problems involve using hand-crafted heuristics that sequentially construct a solution. 
Such heuristics are designed by domain experts and may often be suboptimal due to the hard nature of the problems.
\emph{Reinforcement learning (RL)} proposes a good alternative to automate the search of these heuristics by training an agent in a supervised or self-supervised manner.''

To address the scalability challenge in large-scale CO, \citet{qiu2022dimes} propose an approach called \emph{Differentiable Meta Solver (DIMES)}.
Unlike previous deep reinforcement learning methods, which suffer from costly autoregressive decoding or iterative refinements of discrete solutions, DIMES introduces a compact continuous space for parameterizing the underlying distribution of candidate solutions.
Such a continuous space allows stable REINFORCE-based training and fine-tuning via massively parallel sampling. 

\citet{aironi2024graph} proposed a graph-based neural approach to linear sum assignment problems, which are well-known CO problems with applications in domains such as logistics, robotics, and telecommunications.
In general, obtaining an optimal solution to such problems is computationally infeasible even in small settings, so heuristic algorithms are often used to find near-optimal solutions.
Their paper investigated a general-purpose learning strategy that uses a bipartite graph to describe the problem structure and a message-passing graph neural network model to learn the correct mapping.
The proposed graph-based solver, although sub-optimal, exhibited the highest scalability, compared with other state-of-the-art heuristic approaches.

\citet{georgiev2024neural} note that
``Solving NP-hard/complete combinatorial problems with neural networks is a challenging research area that aims to surpass classical approximate algorithms.
The long-term objective is to outperform hand-designed heuristics for NP-hard/complete problems by learning to generate superior solutions solely from training data.''
The authors proposed leveraging recent advancements in neural algorithmic reasoning to improve learning of CO problems.

\citet{balcan2018learning,balcan2024learning} provide the first sample complexity guarantees for tree search parameter tuning, bounding the number of samples sufficient to ensure that the average performance of tree search over the samples nearly matches its future expected performance on the unknown instance distribution.
\citet{balcan2021sample} prove the first guarantees for learning high-performing cut-selection policies tailored to the instance distribution at hand using samples.
\citet{balcan2022structural} derive sample complexity guarantees for using machine learning to determine which cutting planes to apply during branch-and-cut.

\end{document}